\documentclass{article}

\usepackage[margin=1in]{geometry}

\usepackage{natbib}
\usepackage{paper}

\usepackage{etoolbox}      
\makeatletter
\patchcmd{\thm@space@setup}%
  {\thm@preskip=\topsep}   {\thm@preskip=4pt}{}{}
\patchcmd{\thm@space@setup}%
  {\thm@postskip=\thm@preskip}{\thm@postskip=4pt}{}{}
\makeatother

\usepackage[utf8]{inputenc} 
\usepackage[T1]{fontenc}    
\usepackage{hyperref}       
\usepackage{url}            
\usepackage{booktabs}       
\usepackage{amsfonts}       
\usepackage{nicefrac}       
\usepackage{microtype}      
\usepackage{xcolor}         
\usepackage[capitalize, noabbrev]{cleveref}
\crefname{assumption}{assumption}{assumptions}   
\Crefname{assumption}{Assumption}{Assumptions}   
\usepackage{subcaption}
\usepackage{wrapfig}

\title{Retrieval-Augmented Generation as Noisy In-Context Learning: A Unified Theory and Risk Bounds}

\renewcommand{\and}{\unskip \quad}  

\author{
Yang Guo\thanks{\texttt{
yguo@cs.wisc.edu}. University of Wisconsin-Madison.}
\and
Yutian Yao\thanks{\texttt{
ytao37@wisc.edu}. University of Wisconsin-Madison.}
\and
Yifei Ming\thanks{\texttt{
yifei.ming@salesforce.com}. Salesforce AI Research.}
\and 
Robert D. Nowak\thanks{\texttt{
rdnowak@wisc.edu}. University of Wisconsin-Madison.}
\and
Yingyu Liang\thanks{\texttt{
yingyul@hku.hk}. The University of Hong Kong. \texttt{
yliang@cs.wisc.edu}. University of Wisconsin-Madison.} 
}

\begin{document}

\maketitle

\begin{abstract}
Retrieval-augmented generation (RAG) has seen many empirical successes in recent years by aiding the LLM with external knowledge. However, its theoretical aspect has remained mostly unexplored. In this paper, we propose the first finite-sample generalization bound for RAG in in-context linear regression and derive an exact bias-variance tradeoff. Our framework views the retrieved texts as query-dependent noisy in-context examples and recovers the classical in-context learning (ICL) and standard RAG as the limit cases. Our analysis suggests that an intrinsic ceiling on generalization error exists on RAG as opposed to the ICL. Furthermore, our framework is able to model retrieval both from the training data and from external corpora by introducing uniform and non-uniform RAG noise. In line with our theory, we show the sample efficiency of ICL and RAG empirically with experiments on common QA benchmarks, such as Natural Questions and TriviaQA.

\end{abstract}

\section{Introduction} 

Retrieval-Augmented Generation (RAG) enhances language models by appending retrieved texts to the input, enabling access to information beyond pretraining. It is widely used in open-domain QA, fact-checking, and knowledge-intensive tasks~\citep{Huang2023MakeAnAudioTG,LewisRAG, Ramos2022SmallcapLI,Sarto2022RetrievalAugmentedTF,zhao2024retrievalaugmented}. Retrieval sources typically fall into two categories: (1) \textit{labeled dataset}, such as training dataset itself~\citep{liu2021makes, izacard2022atlasfewshotlearningretrieval,huang2024ravenincontextlearningretrievalaugmented}, and (2) \textit{generic corpora without labels}, such as Wikipedia~\citep{chen2017reading}. Despite its promise, empirical studies show that increasing the number of retrieved passages can degrade performance, especially when irrelevant or redundant texts are included~\citep{levy2025more,levy2024same}. However, the theoretical aspects for understanding of how retrieval affects generalization remain underexplored. 

To study its behavior, we frame RAG as noisy in-context learning (ICL). ICL refers to the ability of language models to adapt given the contextual information without updating model weights~\citep{dong-etal-2024-survey}. Under this view, retrieved RAG examples can act as noisy context and its quality depends on the retrieval. This view has motivated the development of many work in in-context retrieval~\citep{luo2024context,shi2022xricl}, where the goal is to retrieve high-quality demonstrate pairs, which reduces the noise of the retrieval. 

From a theoretical standpoint, RAG becomes tractable when framed as structured in-context learning, where the context consists of fixed format demonstration pairs. Prior ICL work has analyzed this regime under clean, i.i.d. examples~\citep{ahn2023transformers, zhang2024trained}. These assumptions do not hold in RAG, where retrieved examples are noisy, and their noise level tends to be inversely correlated to their relevance. Currently, no theoretical framework has been developed to study RAG under this structured ICL formulation. Although retrieved examples close to the query should, in principle, improve the predictive accuracy, their quantitative contribution remains unknown because RAG introduces these examples only at the test time (absent during pretraining), thus imposing a distribution shift. In this work, we bridge this gap by modeling RAG as noisy ICL, where retrieved examples follow a structured but perturbed distribution. In particular, we model the retrieval noise both under the uniform (same across examples) and non-uniform (inversely correlated with the retrieval relevance). This view allows us to quantify the impact of retrieval noise and derive generalization bounds that depend on the number of in-context and RAG examples, and the retrieval distance from queries.

Our contributions are summarized as follows: 
\begin{itemize}
    \item We propose a theoretical framework for analyzing RAG and provide the first finite sample bounds for in-context linear regression with RAG. Our bounds show that the improvement from RAG shrinks as you add more retrieved examples, and can even flip to hurt performance, giving concrete guidance on when to stop.
    \item Our framework includes ICL and standard RAG as limit cases, and also models retrieved data under different noise regimes, uniform and non-uniform retrieval noise.
    \item We develop new tools for analyzing the query-dependent RAG data, e.g. a derivation of the expectation for 6th-order Gaussian monomial (\cref{lem:6order_gaussian}), which can be useful for future research on RAG.
    \item We conduct experiments for representative models on common QA datasets and demonstrate that early RAG retrieves lie in the uniform noise regime, while later ones shift to non-uniform noise regime, aligning with our theory.
\end{itemize}
\section{Related Work}
\paragraph{Retrieval Augmented Generation}
Retrieval-augmented generation (RAG) has emerged as a widely adopted paradigm for enriching LLMs with external knowledge by prepending retrieved passages to the input context~\citep{LewisRAG,Izacard2020LeveragingPR,Borgeaud2021ImprovingLM}. From a functional perspective, RAG transforms the model’s input distribution by conditioning generation on retrieved textual evidence, often drawn from large-scale corpora via learned or heuristic retrieval mechanisms~\citep{Li2023MakingLL,SFRAIResearch2024,Chen2024BGEMM}. While much of the literature focuses on improving retrieval quality, system performance~\citep{asai2023selfrag,li2024chainofknowledge,xu2024recomp}, and answer reliability~\citep{xiang2024certifiably,xu2024recomp}, the theoretical foundations of RAG remain underexplored. 
\paragraph{In-context Learning (ICL)}
ICL obtains its popularity from the original GPT-3 paper~\citep{brown2020language}, and becomes widely used in LLM applications~\citep{dong-etal-2024-survey,min2021metaicl}. The recent advance in ICL theory~\citep{ahn2023transformers, zhang2024trained, xie2021explanation} provides a rigorous and versatile framework to study transformers and LLMs. People have use this ICL framework to study novel setting, like out-of-distributions tasks~\citep{wang2024can} and test-time training~\citep{gozeten2025test}. People also have also studied the noisy in-context learning from robustness~\citep{cheng2025exploring} and calibration perspectives~\citep{zhao2024noisyicl}, which are different from our setup.

\paragraph{In-context Retrieval}
In-context retrieval~\citep{luo2024context} refers to retrieving a set of query-dependent demonstrations than using fixed set of demonstrations. The label of the demonstration pairs can come from various sources, such as in-domain training set~\citep{izacard2022atlasfewshotlearningretrieval, huang2024ravenincontextlearningretrievalaugmented, ye2023compositional}, cross-domain data~\citep{cheng2023uprise, shi2022xricl}, automatic LLM generation~\citep{zhang2022automatic, li2023mot}, pseudo-labels from unstructured data~\citep{lyu2022z,li2022self}. In our theoretical analysis and experiments, we focus on the simplest in-context retrieval, in-domain retrieval from the training set, as in~\citep{izacard2022atlasfewshotlearningretrieval,huang2024ravenincontextlearningretrievalaugmented}. Note that in-context retrieval is a term developed later and some earlier papers discuss ICL with retrieval as retrieving relevant documents without labels~\citep{ram2023context}.

\section{Problem Setup}

Our problem setup is similar to~\citep{zhang2024trained, garg2022can}, with RAG examples to form the additional in-context examples. It is worth noting that many works focus on ICL at test (inference) time, specifically without parameter updates~\citep{dong2022survey}. Our work adopts the framework of \textit{ICL with warmup}, also known as, \textit{supervised in-context training}. Specifically, we assume that the pretraining data is also formed by in-context examples. Then, during the test time, we formed prompts with in-context examples with additional RAG examples. 

\paragraph{Notations}
We denote $[n] = \{1, \dots, n\}$ for an integer $n \geq 1$. We denote the trace product of two matrices $A,B \in \bbR^{m \times n}$ as $\tr (AB^\top)$.

\paragraph{Pretraining Data}
We consider learning over linear regression data. The training data is a set of prompts. Each prompt is of size $m$: $(\bx_1, y_1, \dots, \bx_m, y_m, \bx_q) \in \mathbb{R}^{d(m+1)+m}$ where $(\bx_1, y_1), \dots, (\bx_m, y_m)$ form the $m$ demonstration pairs. The goal is to predict $\hat{y}_q$ for the query example $\bx_q$ to match the true label $y_q$. The prompt is embedded in the following form: 
\begin{equation}
\PPT_{m} := \begin{pmatrix}
    \bx_1 & \bx_2 & \dots & \bx_m & \bx_q \\ 
    y_1 & y_2 & \dots & y_m & 0 
\end{pmatrix} \in \mathbb{R}^{(d+1) \times (m+1)}
\end{equation}
where $(\bx_1, y_1), \dots, (\bx_m, y_m), (\bx_q, y_q) \iid \calD_{\text{pt}}$ (pt denoting Pretraining). The output follows the linear model: 
\begin{equation}
\begin{aligned}
y_i &= \bx_i^\T \beta_{\text{pt}} + \epsilon_i, \quad \epsilon_i \iid \calN(0, \sigma^2) \quad \text{under} \quad \DPT
\end{aligned}
\end{equation}
where $i \in [m]\cup \{q\}$, $\betatt$ is the weight vector in pretraining, and $\epsilon_i$ is the noise for example $i$.

\paragraph{Inference Data (with RAG)} 
During inference/test time, the test prompt $\bP^{\text{tt+rag}}_{m,n}$ (tt denoting test-time) is formed by $m$ in-context pairs $(\bx_1, y_1), \dots, (\bx_m, y_m)$, $n$ retrieval-augmented pairs $(\bx_1^{\text{rag}}, y_1^{\text{rag}}), \dots,( \bx_n^{\text{rag}}, y^{\text{rag}})$, and the query pair $\bx_q, y_q$.  The test prompt is embedded in the following form:
\begin{equation}
\PTTRAG_{m,n} :=  \begin{pmatrix}
    \bx_1 & \dots & \bx_m &\bx_1^{\text{rag}} & \dots &\bx_n^{\text{rag}} & \bx_q \\ 
    y_1 & \dots & y_m & y_1^{\text{rag}} & \dots &y_n^{\text{rag}} & 0 
\end{pmatrix} \in \mathbb{R}^{(d+1) \times (m+n+1)}
\end{equation}
The input $\bx$ in each in-context or query pair follows the test-time distribution $\calD_{\text{tt}}$, and the label is:
\begin{equation}
\begin{aligned}
y_i =  \bx_i^\T \beta_{\text{tt}} + \epsilon_i, \quad \epsilon_i \iid  \calN(0, \sigma^2) \quad \text{under} \quad \calD_{\text{tt}}
\end{aligned}
\end{equation}
where $i \in [m]\cup\{q\}$, $\epsilon_i$ is the noise of example $i$, and $\betatt$ is the weight vector during test time. 
The input $\bx$ in each RAG pair follows the corresponding RAG distribution $\calD_{\text{rag}}(\bx_q)$:  
assume the RAG query $\bx_i^{\text{rag}} = \bx_q + \br_i$ is generated around the query example $\bx_q$, where $\br_i$ is the offset. The label in the RAG example is given by: 
\begin{equation}
\begin{aligned}
 \quad y_i^{\text{rag}} =  (\bx_i^{\text{rag}})^\T& \beta_{\text{{tt}}} + \epsilon_i^{\text{rag}} , \quad \epsilon^{\text{rag}}_i \iid  \calN(0, \sigma^2_{\text{rag}, i}) \quad \text{under} \quad \calD_{\text{rag}}(\bx_q) 
\end{aligned}
\end{equation}
where $i\in [n]$, $\epsilon^{\text{rag}}_i$ is the noise of the $i$-th RAG example. 

For the compactness of writing, we define the following matrices and vectors: 
\[
\Xicl := [\bx_1^\T ; \dots ;\bx_m^\T], \; \Xrag := [(\bx_1^{\text{rag}})^\T ; \dots ;(\bx_n^{\text{rag}})^\T], \; \yicl := [y_1; \dots ; y_m],  \; \yrag := [y_1^{\text{rag}}; \dots ; y_n^\text{rag}],\; 
\]
\[
\bepsicl := [\epsilon_1;\dots ;\epsilon_m],\; \bepsrag := [\epsilon_1^\text{rag}; \dots ; \epsilon_n^\text{rag}], \; \br = [\br_i^\T; \dots ; \br_n^\T]
\]
\[
\bX=\begin{bmatrix} \Xicl \\ \Xrag\end{bmatrix} \in \mathbb{R}^{(m+n) \times d}, \; \Xrag = \begin{bmatrix}
    \bx_q+\br_1\\\vdots\\\bx_q+\br_n
\end{bmatrix} \in \mathbb{R}^{n \times d}, \; \by = \begin{bmatrix}\yicl \\ \yrag
\end{bmatrix} \in \mathbbm{R}^{m+n}, \; \beps = \begin{bmatrix}
    \bepsicl \\ \bepsrag
\end{bmatrix} \in \mathbbm{R}^{m+n}
\]

\paragraph{Training and Testing}
We let $\bW$ be the model parameters, and $F$ be the model.
Given an input prompt $\bP_{m}^{\text{pt}}$ with demonstration pairs, the model predicts $\hat{y}_q:= F(\bP_{m}^{\text{pt}}; \bW)$. As a common practice in theoretical studies of LLM for feasible analysis, we use the MSE loss as the evaluation metrics~\citep{zhang2024trained,ahn2023transformers,xie2021explanation}. Then, the population loss on the pretraining data is:
\begin{equation}
\calL_{\text{pt}}(\bW) := \Exp_{(\bx_1, y_1), \dots, (\bx_m, y_m) (\bx_q,y_q)\sim \DPT} \left[ \left(y_q - F\left(\bP_{m}^{\text{pt}}; \bW\right)\right)^2\right]
\end{equation}
Its minimizer is denoted as: 
\begin{equation}
\bar{\bW}^* := \min_{\bW} \calL_{\text{pt}}(\bW).
\end{equation}

To apply the pretrained $\bW^*$ from the pretraining context size of $m$ to the test-time context size of $m+n$, we will need to scale it properly (see~\cref{lem:adaptW}) and use 
\begin{equation}\label{eq:adapt_W}
\bW^* = \frac{m}{m+n}\bar{\bW}^*.
\end{equation}

During the test time we evaluate the population loss over the test prompt with RAG examples $\bP_{m,n}^{\text{tt+rag}}$:
\begin{equation}
\calL_{\text{tt+rag}}(\bW) := \Exp_{\substack{(\bx_1, y_1), \dots, (\bx_m, y_m) (\bx_q,y_q)\sim \DTT \\ (\bx^{\text{rag}}_1, y^{\text{rag}}_1), \dots, (\bx^{\text{rag}}_n, y^{\text{rag}}_n) \sim \DRAG(\bx_q)}} \left[ \left(y_q - F\left(\bP_{m,n}^{\text{tt+rag}}; \bW\right)\right)^2\right] 
\end{equation}

\paragraph{Model Architecture}
We study the single-layer linear self-attention model (LSA) as the framework for theoretical analysis, similar to many existing studies (e.g.,~\citep{ahn2023transformers, zhang2024trained}). The prediction of the model $F$ on a prompt $\bP$ with query $\bx_q$ is: 
\begin{equation}\label{eq:prediction_form}
\hat{y}_q := F(\bP) = [\bP\bW_Q \bW_K^\T \bP^\T \bP \bW_V]_{m+n+1, d+1}=  \bx_q^\T \bW \bX^\T \by 
\end{equation}
where the query, key, and value matrices $\bW_Q, \bW, \bW_V \in \mathbb{R}^{(d+1)\times(d+1)}$ are parameterized by $\bW$ in the follow way: 

\[
\bW_Q \bW_K^\T = \begin{bmatrix}
    \bW & \bzero_{d\times 1} \\ 
    \bzero_{1\times d} & 0
\end{bmatrix}, \quad \bW_V = \begin{bmatrix}
    \bzero_{d\times d} & \bzero_{d\times 1} \\ 
    \bzero_{1\times d} & 1 
\end{bmatrix}
\]
We note that this parameterization is commonly used in the previous works~\citep{ahn2023transformers, zhang2024trained}, and is shown to capture the key properties of in-context learning. Furthermore, \citep{ahn2023transformers} shows that the formulation is the optimum converged from pretraining on Gaussian data.

\section{Theoretical Analysis: Generalization Bound for RAG} 

To study test-time error and sample complexity in in-context linear regression with RAG examples, we consider two noise regimes: \textbf{uniform retrieval noise} and \textbf{non-uniform retrieval noise}. Uniform retrieval noise assumes the RAG noise $\epsilon_{i}^\rag$ for each example $i$ is i.i.d. Since its variance is distance-agnostic, it can model a scenario of retrieval where the noise is prevailing across data points. Non-uniform retrieval noise assumes either the variance or the label-corruption probability grows with the variance of retrieval vector --- e.g. $\sigma_{\rag,i}^2$ increases with $\delta_i^2$ or probability of making mistakes increases with $\delta_i^2$. This captures retrieval from datasets where near neighbors often supply the right signal while far ones are potentially noisy or misleading. Because the noise spectrum is now heavy-tailed, adding more RAG examples past a threshold could yield diminishing benefits for RAG examples and even become counter-productive. Framing RAG through these two lenses allows precise clarification about when extra retrieved examples will pay off, and when they will hit the intrinsic ceiling and more retrieved examples don't help anymore. These are well corroborated by our experimental results on real data (see Section~\ref{sec:exp}).

First, we introduce the key data assumptions. 

\begin{assumption}[Gaussian Retrieval Offset]\label{ass:gaussian_offset}
We assume the retrieval offset $\br_i, \; \forall i \in [n]$ to follow a Gaussian distribution:
$
\br_i \stackrel{\text { i.i.d. }}{\sim} \mathcal{N}\left(0, \delta_i^2 I_d\right).
$
\end{assumption}
The key property that we want to control for RAG examples is its distance from the query points $\bx_q$. However, modeling the queried example directly through the retrieval distance leads to complicated theoretical analysis. Here, we note that the retrieval distance $\|\br_i\|_2$ converges to a distribution concentrated in an $\calO(\delta_i\sqrt{d })$ ball around the query with respect to $d$~\citep{cover1967nearest}. Thus, controlling the variance of the retrieval offset can alternatively control the retrieval distance. And we make the following additional data assumptions.
\begin{assumption}[Data Assumption]\label{ass:data}
We assume the data follows the following: 
\begin{enumerate}\setlength\itemsep{0em}
    \item \textsc{Pretraining Examples ($\DPT$)}. For a pretraining prompt of length $m+1$ and for all $i \in [m]\cup\{q\}$, we assume
          $\bx_i\!\iid\!\mathcal{N}(0,I)$,\;
          $\epsilon_i\!\iid\!\mathcal{N}(0,\sigma^{2})$,\;
          $\beta_{\mathrm{pt}}\!\sim\!\mathcal{N}(0,I)$.
    \item \textsc{Test Time Examples ($\DTT$)}.
           For a test-time prompt of length $m+n+1$ and for all $i \in [m]\cup\{q\}$, we assume $\bx_i \!\iid\!\mathcal{N}(0,I)$,\;
          $\epsilon_i\!\iid\!\mathcal{N}(0,\sigma^{2})$,\;
          $\beta_{\mathrm{tt}}\!\sim\!\mathcal{N}(0,I)$.
    \item \textsc{Test-Time RAG Examples ($\DRAG(\bx_q)$)}.  
          For a test-time prompt of length $m+n+1$ and for all $i \in [m+1, \dots, m+n]$, we assume $\bx_i^{\rag}\!\iid\!\mathcal{N}(0,I)$,\;
          $\epsilon_i^{\rag}\!\sim\!\mathcal{N}(0,\sigma_{\rag, i}^{2})$,  
          and the same $\beta_{\mathrm{tt}}$ as (2). 
\end{enumerate}
\end{assumption}
Here, we assume the isotropic Gaussian property for the input, noise and the weight vector, a common assumption made in ICL theory~\citep{ahn2023transformers,gozeten2025test} for simple yet meaningful analysis. 

\begin{AIbox}{Overview of the Key Results}
\begin{itemize}[leftmargin=0em,noitemsep]
    \item (\textit{Uniform Noise}) RAG examples are as effective as ICL examples in reducing the variance-induced err but ineffective at reducing the bias-induced err, causing a loss plateau for $n\rightarrow \infty$,
    \item (\textit{Non-Uniform Noise}) RAG could improve the variance-induced error up to a finite $n$ at a cost of increasing bias-induced error.
\end{itemize}
\end{AIbox}

\paragraph{Roadmap}
Under these assumptions and uniform retrieval noise, we will first derive the population loss of RAG, $\calL_{\mathrm{tt+rag}}(\bW)$, for general $\bW$ as in~\cref{thm:pop-loss-rag}, analyze its finite sample complexity under the optimal pretrained weight $\bW^*$ as in~\cref{prop:rag_loss_nonasymptotic} and derive an optimal number of RAG examples of $n^*$ for a given number of ICL examples $m$ as in~\cref{prop:optimal_n}. These discussions leads to our first key result. Then, under the non-uniform retrieval noise, we will prove the sample complexity under the distance-proportional noise (\cref{thm: bound_proportion_noise}) and distance-weighted mixture noise (\cref{thm:bound_probablistic_noise}), and obtain our second key results above.

\subsection{Uniform Retrieval Noise}\label{sec:uniform_noise}
\begin{assumption}[Uniform Retrieval Noise]\label{ass:uniform_noise}
The RAG noise $\beps_{\mathrm{rag}}$ shares the same Gaussian distribution with variance $\sigma^2_{\mathrm{rag}}$, i.e. $\forall i \in [m+1, \dots,m+n]$,\; 
$\sigma_{\mathrm{rag}, i}^2 = \sigma_{\mathrm{rag}}^2 $.
\end{assumption}
First, we present the assumption for uniform retrieval noise. In other words, all RAG examples are as helpful, and its improvement on the actual prediction is determined by the retrieval distance.

\begin{theorem}[Population Loss for ICL with RAG Examples]\label{thm:pop-loss-rag}
Under Assumption~\ref{ass:gaussian_offset}, \ref{ass:data}, \ref{ass:uniform_noise}, the population loss of the linear self-attention predictor
$\hat{y}_q=\bx_q^\T\boldsymbol W\bX^\T\by$
satisfies
\begin{equation}
\mathcal{L}_{\mathrm{tt+rag}}(\bW)=\underbrace{\Exp\left(\Exp\left(\hat{y}_q\right)-\hat{y}_q\right)^2}_{:=\operatorname{err}_{\mathrm{variance}}(\bW)}+\underbrace{\Exp\left(\Exp\left(\hat{y}_q\right)-\Exp\left(y_q\right)\right)^2}_{:=\operatorname{err}_{\mathrm{bias}}(\bW)}+\underbrace{\sigma^2}_{\text {irreducible noise}} \quad \text{, and specifically, }
\end{equation}
\begin{align*}
\err_{\mathrm{variance}}(\bW) &= \left[m\sigma^2 + \left( 1  
+\delta^2\right)n\sigma_{\mathrm{rag}}^2\right]\tr(\bW^\T\bW) + n \sigma_{\mathrm{rag}}^2 \tr(\bW^2) + n\sigma_{\mathrm{rag}}^2 \tr(\bW)^2 \\ 
\err_{\mathrm{bias}}(\bW)&= \betatt^\T \left[I -  (n \delta^2 + 2n + m) (\bW+ \bW^\T) - 2n\tr(\bW)I + M_4\right]\betatt \\ 
&=\betatt^\T \left[I -  (n \delta^2 + 2n + m) (\bW+ \bW^\T) - 2n\tr(\bW)I  \right. \\ 
&\quad +\left[n^2\left(2+\delta^2\right)+n\left(m+\delta^2\right)\right]\left(\bW^2+\left(\bW^2\right)^{\top}\right) +2n(n+\delta^2) \bW \bW^{\top} \\ 
& \quad + \left[m^2+m+m n\left(2+2 \delta^2\right)+n^2\left(2+2 \delta^2+\delta^4\right)+n\left(2 \delta^2+\delta^4\right)\right] \bW^{\top} \bW \\ 
&\quad+\left[n^2\left(2+\delta^2\right)+n\left(m+\delta^2\right)\right]\left(\tr(\bW)\left(\bW+\bW^{\top}\right)\right) \\ 
&\left. \quad +\left[n^2+n \delta^2\right]\left(\tr(\bW)^2+\tr\left(\bW^2\right)\right) I +\left[m+n^2+n\left(2 \delta^2+\delta^4\right)\right]\tr\left(\bW^{\top} \bW\right) I  \right] \betatt
\end{align*}

\end{theorem}
Here, we derive the exact bias-variance decomposition for ICL with RAG. The first line is the variance-induced error formed by a weighted sum of noise from ICL examples and RAG examples. Because of the implicit scaling of $\bW$ as discussed in~\cref{lem:adaptW}, the second order term in $\bW$ will introduce an additional weight scaling of $\frac{m^2}{(m+n)^2}$ when adapting from the weight learned on $m$ size context to $m+n$ size context. Thus, larger $n$ will let $\err_\var(\bW) \rightarrow 0$, and the convergence rate is affected by $\delta^2$. Larger retrieval distance leads to a slower convergence. The bias-induced error is composed of all possible monomials of $\bW$ up to the $2$nd-order with $\tr$ operation. The complex dependency on $m,n,\delta^2, d$ requires additional assumptions on $\bW$ to further interpret. As a sanity check, when $n=0$ (ICL-only), this decomposition can exactly recover loss as in Lemma B.2 in~\citep{gozeten2025test}. 

As a proof sketch, we first compute $\err_\var(\bW) = \Exp (\bx_q^\T \bW \bX^\T \beps)^2$ by splitting the calculation for ICL and RAG examples based on $\bX$. Then, we compute $\err_\bias(\bW) = \Exp [( \bx^\T_q(I-\bW \bX^\T\bX ) \betatt )^2]$. The main technical challenge lies in the dependency of $\bX_{\rag}$ on $\bx_q$, and $\err_\bias$ has a 6th-order dependency on $\bx_q$ (2 from $\bx_q$ and 4 from $\bX$). As shown in~\cref{lem:6order_gaussian}, $\Exp \left[ \bx_q \bx_q^{\top} A \bx_q \bx_q^{\top} B \bx_q \bx_q^{\top}\right]$ gives 15 new terms that include all the second order monomials of $\bW$ with $\tr$. The calculation requires multiple careful applications of Isserlis' theorem~\citep{isserlis1918formula}, and the full proof can be seen in~\cref{sec:app_rag_proof}. It is possible to prove this theorem for a design matrix with non-isotropic covariance, but computing the expectation of the 6th-order Gaussian monomial is more complicated. 

Here, we present the finite sample bound for pretrained $\bW^*$ for better interpretation.
\begin{proposition}[Finite Sample Generalization Bound]\label{prop:rag_loss_nonasymptotic} Under Assumption~\ref{ass:gaussian_offset}, \ref{ass:data}, \ref{ass:uniform_noise}, if $\delta^2 \ll 1$, 
\[
\calL_{\mathrm{tt+rag}}(\bW^*) =\calO \left(\sigma^2 + \underbrace{\frac{dm}{(m+n)^2}\sigma^2 + \frac{d^2n}{(m+n)^2}\sigma_{\mathrm{rag}}^2}_{\err_\var(\bW^*)} + \underbrace{\|\betatt\|_2^2 \left[ \frac{d}{m} + d^2\left(\frac{n}{m+n}\right)^2 \right]}_{\err_\bias(\bW^*)}\right)
\]
\begin{equation}
\begin{aligned}
\err_{\mathrm{variance}}(\bW^*)
&= \begin{cases}
 \calO(\frac{d}{m}\sigma^2 + \frac{d^2}{m^2}\sigma_{\mathrm{rag}}^2 ) = \calO\left(\frac{1}{m}\right) & \text{ $m \rightarrow \infty$, $n$ fixed.}
 \\ 
\calO(\frac{d}{n^2}\sigma^2 + \frac{d^2}{n}\sigma_{\mathrm{rag}}^2) =  \calO\left(\frac{1}{n}\right)& \text{ $n \rightarrow \infty$, $m$ fixed} \\ 
\calO(\frac{d}{m}\sigma^2 + \frac{d^2}{m}\sigma_{\mathrm{rag}}^2) = \calO \left(\frac{1}{m}\right) & \text{ $m,n \rightarrow \infty$, $n = \Theta(m)$}    
\end{cases}
\end{aligned}
\end{equation}

\begin{equation}
\begin{aligned}
\err_{\mathrm{bias}}(\bW^*)
&= \begin{cases}
\calO\left(\|\betatt\|_2^2 \frac{d}{m}  \right) & \text{if $m\rightarrow \infty$, $n$ is fixed} \\ 
 \calO\left(\|\betatt\|_2^2  d^2 \right)  = C_1 &\text{if $n\rightarrow \infty$, $m$ is fixed} \\ 
\calO\left(\|\betatt\|_2^2 \left( \frac{d}{m}+ d^2\right) \right) = C_2 + \calO(\|\betatt\|_2^2\frac{d}{m}) &\text{if $m\rightarrow \infty$, $n = \Theta(m)$} \\ 
\end{cases}
\end{aligned}
\end{equation}
\end{proposition}
Here, we assume $\delta^2\ll 1$ as the test time example $\bx_i$ has only a variance of $I$, and it is unrealistic to assume a higher retrieval variance than the input variance. On the limit case where $m \rightarrow \infty$ and $n$ are fixed, we observe that both variance-induced and bias-induced error decay at a rate of $\calO \left(1/m\right)$, matching the results from the existing paper~\citep{ahn2023transformers, zhang2024trained}. When $n \rightarrow \infty$, the variance-induced error decays as $\calO\left(1/n\right)$ matching the $\calO(1/m)$ rate. However, introducing the RAG is ineffective at reducing the bias-induced error. Even when $m\rightarrow \infty$, increasing $n$ will cause a loss plateau. 

This effect can be explained by the underlying adaptive ability of transformers. In an online learning setup, we could always use the mean of the queried data as the prediction. However, in the LSA setup, the pretrained $\bW^*$ serves as a proxy for $\Exp^{-1} (\bX^\T \bX)$. In order to retain the adaptivity to the entire distribution of $\betatt$, we cannot use the optimal linear classifier $(\bX^\T \bX)^{-1}\bX^\T\by$ or use the mean of the retrieved examples ad hoc. At the test stage, $\bX_\rag$ only appears in $\bX^\T \by$ and not in $\bW^*$. The difference between $\calD_\rag(\bx_q)$ and $\calD_\text{tt}$ directly leads to the increase of variance worsened by the increase of $n$. See full proof in~\cref{sec:app_rag_proof}. Now, a natural question is whether we can find a balance of variance and bias and obtain an optimal RAG example size $n^*$.
\begin{proposition}\label{prop:optimal_n}
Under Assumption~\ref{ass:gaussian_offset},\ref{ass:data},\ref{ass:uniform_noise}, $\delta^2 \ll 1$, and reasonable choice of $\sigma^2, \sigma_{\mathrm{rag}}^2$ ($\sigma^2, \sigma_{\mathrm{rag}}^2 \ll \|\betatt\|_2^2$), the optimal $n^*$ that minimizes the RAG loss follows: 
\begin{equation}
n^* = \calO\left(\frac{m \left(d^2 \|\betatt\|_2^2 + d \sigma^2- d^2 \sigma_{\mathrm{rag}}^2\right)}{m d^2 \|\betatt\|_2^2 - d^2\sigma_{\mathrm{rag}}^2}\right) = \calO\left(\frac{d\|\betatt\|_2^2 + \sigma^2-d\sigma_\rag^2 }{d\|\betatt\|_2^2}\right)
\end{equation}
and the improvement on loss from picking the optimal $n^*$ over $n =0$ is given as:
\begin{equation}
\calL_{\mathrm{tt+rag}}(\bW^*)|_{n=0} - \calL_{\mathrm{tt+rag}}(\bW^*)|_{n=n^*} = \calO\left(\frac{1}{m^2}\right)
\end{equation}
\end{proposition}
In fact, the optimal $n^*$ does not scale with $m$ omitting the lower-order terms. Note that for $\|\betatt\|_2^2 = \calO(1)$, $\|\betatt\|_2^2$ will dominate the numerator for reasonable choices of $\sigma^2$ and $\sigma^2_\rag$. A larger ICL noise $\sigma^2$ leads to a larger $n^*$, i.e. requiring more RAG examples to compensate for the loss. A larger RAG noise $\sigma^2_\rag$ leads to a smaller $n^*$, i.e. less efficiency on RAG examples. And the improvement converges at $\calO(\frac{1}{m^2})$, diminishing for large $m$. See the full proof in~\cref{sec:app_rag_proof}. Several empirical works also observe a performance drop when increasing the number of retrieved examples~\citep{wang2024leave,levy2025more}.  

\subsection{Non-Uniform Retrieval Noise}

The uniform‐noise setup in~\cref{sec:uniform_noise} relies on a clean retrieval pool, so we could keep the variance $\sigma_{\mathrm{rag}}^{2}$ fixed. In open-domain retrieval, this assumption could collapse: many retrieved examples could contain no answer or even a wrong answer. Empirically, people have observed that passages that are closer to the query vector $\boldsymbol{x}_q$ are more likely~\citep{yang2020retriever, yoran2023making, lewis2020retrieval} to contain the correct label. We want to theoretically investigate if the following hypothesis still holds: 
\begin{center}
\fbox{Closer to query $\bx_q$ $\implies$ \textit{more likely} to contain \textit{correct} answer.}
\end{center}

\subsubsection{Distance-Proportional Noise (DPN)}
We first investigate the scenario where the retrieval noise is proportional to the retrieval distance. Since the ICL analysis only applies to the mean-squared error loss, we study the effect of RAG under DPN on the correctness of the predictions.
\begin{assumption}[Distance-Proportional Noise]\label{ass:proportional}
There exists a constant $\gamma_1>0$ such that, for every retrieved sample $i$,
$
\sigma_{\mathrm{rag}, i}^2=\gamma_1 \sigma^2 \delta_i^2,
$
i.e. the RAG noise variance grows linearly with the variance $\delta_i^2$ that governs the retrieval distance.
\end{assumption}
Under the new data assumption, we denote the corresponding RAG loss, bias-induced error, and variance-induced error for $\bW$ to be $\hat{\calL}_{\mathrm{tt+rag}}(\bW)$, $\hat{\err}_{\mathrm{bias}}(\bW)$, and $\hat{\err}_{\mathrm{variance}}(\bW)$.

\begin{theorem}[Finite Sample RAG Generalization Bound under DPN]\label{thm: bound_proportion_noise}
Under Assumption~\ref{ass:gaussian_offset}, \ref{ass:data}, \ref{ass:proportional}, the population loss is given as:
\[
\hat{\err}_{\mathrm{variance}}(\bW) = m\sigma^2 \tr(\bW^\T \bW) + \sum_{i=1}^n \gamma_1 \delta_i^2 [(1+\delta_i^2) \tr(
\bW^\T \bW
)+ \tr(\bW^2) + \tr(\bW)^2] 
\]

If the variance of the retrieval distance follows power law, i.e. $\exists \gamma_2 > 0, q \geq 0$ s.t. $ \delta_i^2 = \gamma_2 i^q $, then
\begin{equation}
\begin{aligned}
\hat{\err}_{\mathrm{bias}}(\bW^*)&=
\calO \left(\err_\bias(\bW^*) + \|\betatt\|_2^2 \left[\frac{ dn^{2q+1} + n^{2q+2}}{(m+n)^2}\right]\right)
\end{aligned}
\end{equation}

and 
\begin{equation}
\begin{aligned}
\hat{\err}_{\mathrm{variance}}(\bW^*)&= \calO\left(
\frac{dm\sigma^2 + d(n^{2q+1})\sigma^2}{(m+n)^2}
\right) = \begin{cases}
\calO\left(dn^{2q-1}\sigma^2\right)  & \text{ if $n\rightarrow \infty$, $q \leq 1/2$ } \\ 
\text{diverges} & \text{ if $n\rightarrow \infty$, $q>1/2$} \\ 
\end{cases} 
\end{aligned}
\end{equation}
\end{theorem}
Here, we derive the sample complexity under DPN. A second order dependency on $\delta_i^2$ shows up in both the variance-induced and bias-induced error (exact form seen in~\cref{sec:app_rag_proof}). Thus, the $\delta_i^2$-involved constant will dominate the other constants. Specifically, it even leads to divergence for $q > 1/2$ for the variance-induced error and $q > 0$ for the bias-induced error.

\subsubsection{Distance-Weighted Mixture Noise}
In this section, we discuss the scenario where further RAG examples are less likely to contain the correct answers. We use a pair of large and small noises to model the correct/incorrect examples.
\begin{assumption}[Distance-Weighted Mixture Noise]\label{ass:probablistic}
We assume that the RAG noise is formed by a mixture of small and large noise:
\[
y(\bx_{\mathrm{rag}}) = \begin{cases}
    f(\bx_{\mathrm{rag}, i}) + \epsilon_{s}  &   \text{w.p. $p_i$} \\
   f(\bx_{\mathrm{rag}, i}) + \epsilon_{l} &   \text{w.p. $1-p_i$}
\end{cases}
\]
where $\epsilon_s \sim \calN(0, c_s\sigma^2)$ corresponds to the small noise and $\epsilon_l \sim \calN(0,c_l\sigma^2)$ corresponds to the large noise, with $c_l \geq c_s \geq 0$. The probability of sampling small noise $p_i$ follows an inverse power law of the variance of the retrieval distance, i.e. $p_i =  (1+\delta_i^2)^{-\tilde{q}}, \, \tilde{q} \geq 0 $.
\end{assumption}
Here, we choose the sampling probability (of small noise) $p_i$ to follow a polynomial decay and the constant 1 here is to ensure $p_i =0$ when $\delta_i^2 = 0$. Under the new data assumption, we denote the corresponding RAG loss, bias-induced error, and variance-induced error for $\bW$ to be $\tilde{\calL}_{\mathrm{tt+rag}}(\bW)$, $\tilde{\err}_{\mathrm{bias}}(\bW)$, and $\tilde{\err}_{\mathrm{variance}}(\bW)$.
\begin{theorem}[Finite Sample RAG Bound under Distance-Weighted Mixture Noise]\label{thm:bound_probablistic_noise}
Under Assumption~\ref{ass:gaussian_offset}, \ref{ass:data}, \ref{ass:probablistic}, then $\tilde{\err}_\bias(\bW) = \hat{\err}_\bias(\bW)$, and
\[
\hat{\err}_{\mathrm{variance}}(\bW) = m\sigma^2 \tr(\bW^\T \bW) + \sum_{i=1}^n \left(p_i\sigma_s^2  + (1-p_i)\sigma_l^2 \right) [(1+\delta_i^2) \tr(
\bW^\T \bW
)+ \tr(\bW^2) + \tr(\bW)^2] 
\]
If the variance of the retrieval distance follows power law, i.e. $\exists \gamma_2 > 0, q \geq 0$ s.t. $ \delta_i^2 = \gamma_2 i^q $, then:

\begin{equation}
\begin{aligned}
\tilde{\err}_{\mathrm{variance}}(\bW^*)  
& = \begin{cases}
\calO\left( c_l dn^{q-1}\sigma^2 - (c_l-c_s) \sigma^2 d n^{q-1-q \tilde{q}} \right) & \text{ if $n\rightarrow \infty$, $q \leq 1$} \\ 
\text{diverges} & \text{ if $n\rightarrow \infty$, $q>1$} \\  
\end{cases}
\end{aligned}
\end{equation}
\end{theorem}
The bias-induced error here is the same as in DPN, since we assume a polynomial dependency for $\delta_i^2$ on $i$ in both setting and the bias-induced error is independent of the variance of noise. Even though the variance of small/large noise are bounded, the dependency on the retrieval distance leads to the divergence at large $q$ ($q > 1$). The large prediction noise will dominate the variance-induced error, but a larger gap between large and small noise ($c_l - c_s$) can mitigate the error by a ratio of $\calO(n^{-q\tilde{q}})$. That is, the smaller $q$ and $\tilde{q}$ are, the lower the error. 

We note that the uniform noise scenario can also admit the mixture noise model by taking a constant $p_i, \; \forall i$, resulting in a form similar to the standard uniform retrieval noise in~\cref{prop:rag_loss_nonasymptotic}.

\section{Experiments} \label{sec:exp}
We investigate the effect of RAG focusing on the following questions: (\textbf{Q1}) Whether RAG data outperform randomly sampled in-context examples? (\textbf{Q2}) What are the impacts of the RAG examples from training data and RAG passages from external corpora? (\textbf{Q3}) With a fixed budget of example numbers, what is the effect of varying the ratio between the two types of RAG data? 
Our experiments provide the following findings: (\textbf{A1}) RAG data lead to better performance than in-context ones under different data budgets. (\textbf{A2}) Interestingly, the first few RAG training examples significantly improve performance, but later ones are harmful, because the first few are highly relevant but later ones are noise rather than signal. 
In contrast, RAG passages from external corpora can slowly but monotonically improve the performance, because external corpora are large enough to provide noisy but still relevant data. These are captured by different noise models in our theory. \textbf{(A3)} The performance is not monotonic with the ratio, and the sweet spot depends on the data/model. 

\paragraph{Setup}
For Natural Questions (NQ), the retrieval index is constructed from the December 2018 Wikipedia dump. For TriviaQA, we use the December 2021 version. To accommodate hardware limitations, we randomly subsample 10\% of the full index for both datasets. This reduces retrieval cost and memory usage, allowing all experiments to be conducted on a single NVIDIA A100 or L40 GPU.

We use representative models \textbf{ATLAS} \cite{izacard2022atlasfewshotlearningretrieval} and \textbf{RAVEN} \cite{huang2024ravenincontextlearningretrievalaugmented} on two standard open-domain question answering benchmarks \textbf{Natural Questions (NQ)} \cite{kwiatkowski-etal-2019-natural} and \textbf{TriviaQA} \cite{joshi-etal-2017-triviaqa}.
For evaluation, the context consists of $m$ in-context examples, and $n$ RAG data points (including $n_1$ RAG examples from the training data and $n_2$ RAG passages from external corpora like Wikipedia, so $n=n_1 + n_2$). We choose different $m,n_1,n_2$'s for our study purpose and report the standard exact match (EM) accuracy on 1000 random samples from the test set. 

\begin{figure}[h!]
  \centering
  \begin{subfigure}[b]{0.48\textwidth}
    \centering
    \includegraphics[width=\linewidth]{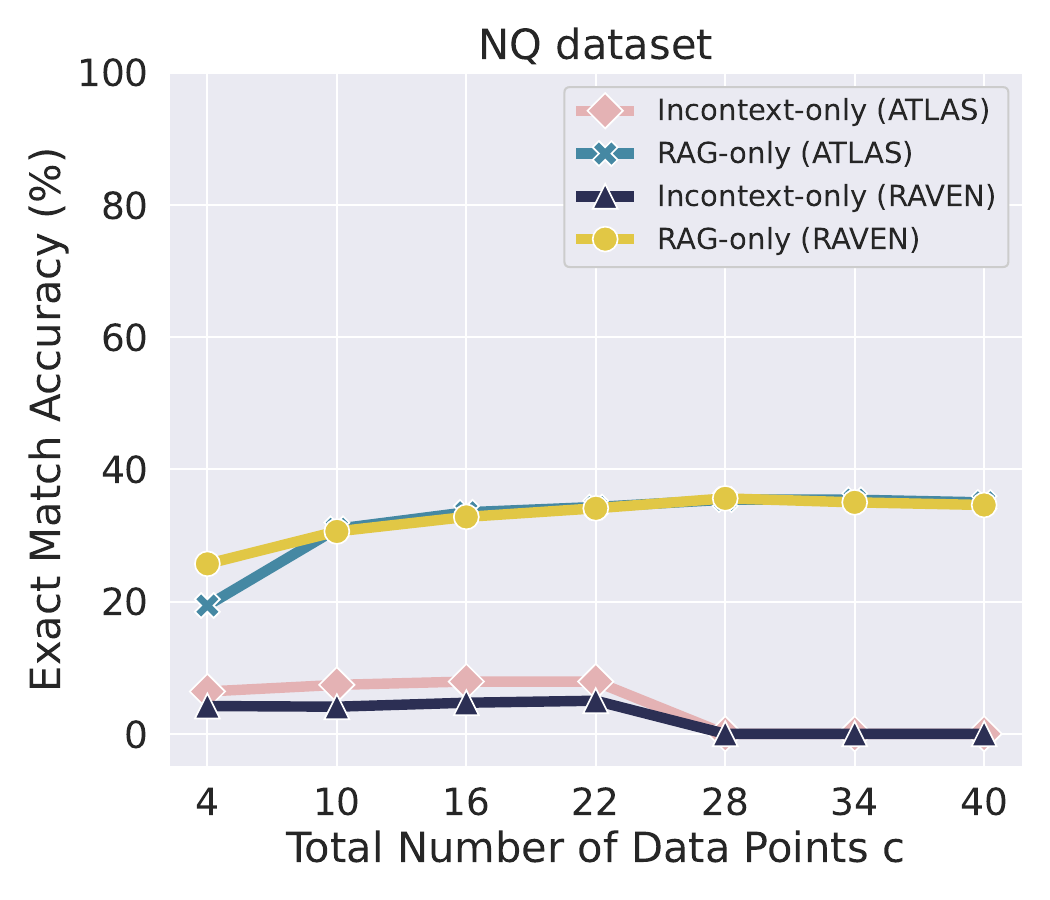}
    \label{fig:exp1_1}
  \end{subfigure}
  \hfill
  \begin{subfigure}[b]{0.48\textwidth}
    \centering
    \includegraphics[width=\linewidth]{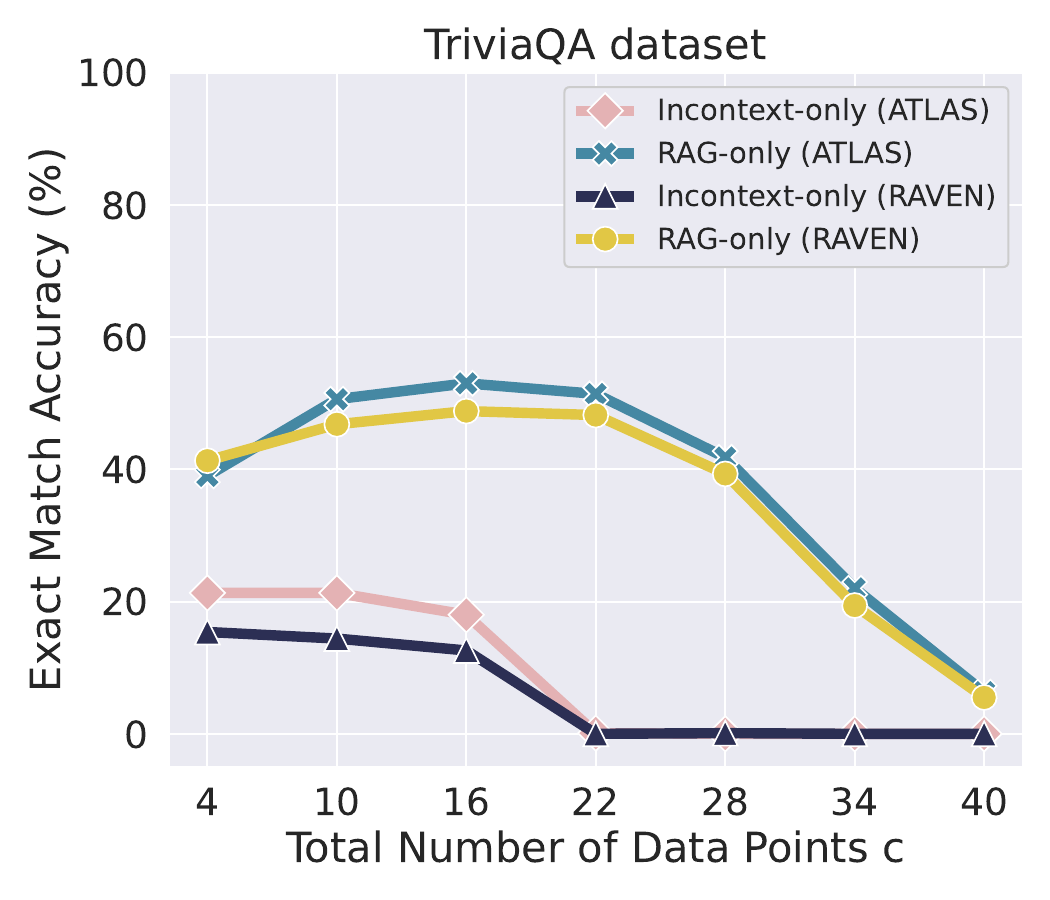}
    \label{fig:exp1_2}
  \end{subfigure}
  \caption{We compare performance between the RAG-only ($c=m$) versus in-context-only methods ($c=n_1+n_2, n_1 = n_2$), where $c$ is the total number of data, $n_1$ refers to retrieved examples and $n_2$ to passages. }\label{fig:rag_ablation1}
\end{figure}

\paragraph{RAG v.s.\ In-Context}
For a budget $c$, we compare using RAG only ($m=0, n_1=n_2=c/2$) and in-context examples only ($m=c, n=0$). The results in Figure~\ref{fig:rag_ablation1} show that RAG consistently outperforms in-context examples, as RAG provides query-relevant data with more signals to address the query, consistent with our analysis.

\begin{figure}[h!]
  \begin{subfigure}[b]{0.48\textwidth}
    \centering
    \includegraphics[width=\linewidth]{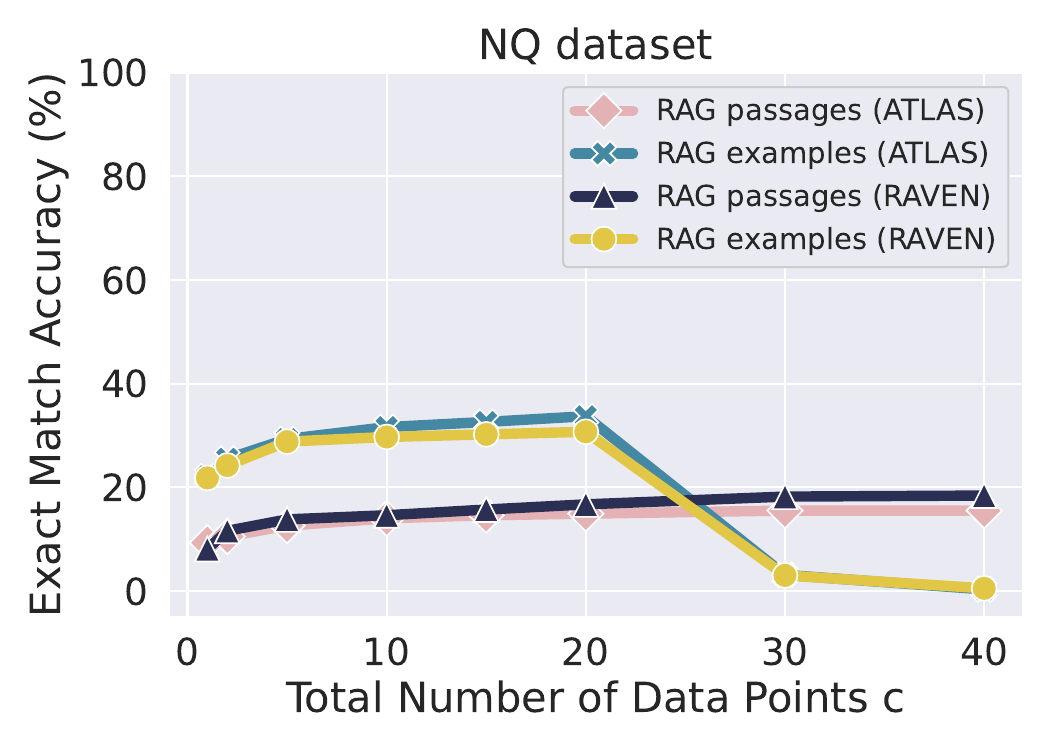}
    \label{fig:exp2_1}
  \end{subfigure}
  \hfill
  \begin{subfigure}[b]{0.48\textwidth}
    \centering
    \includegraphics[width=\linewidth]{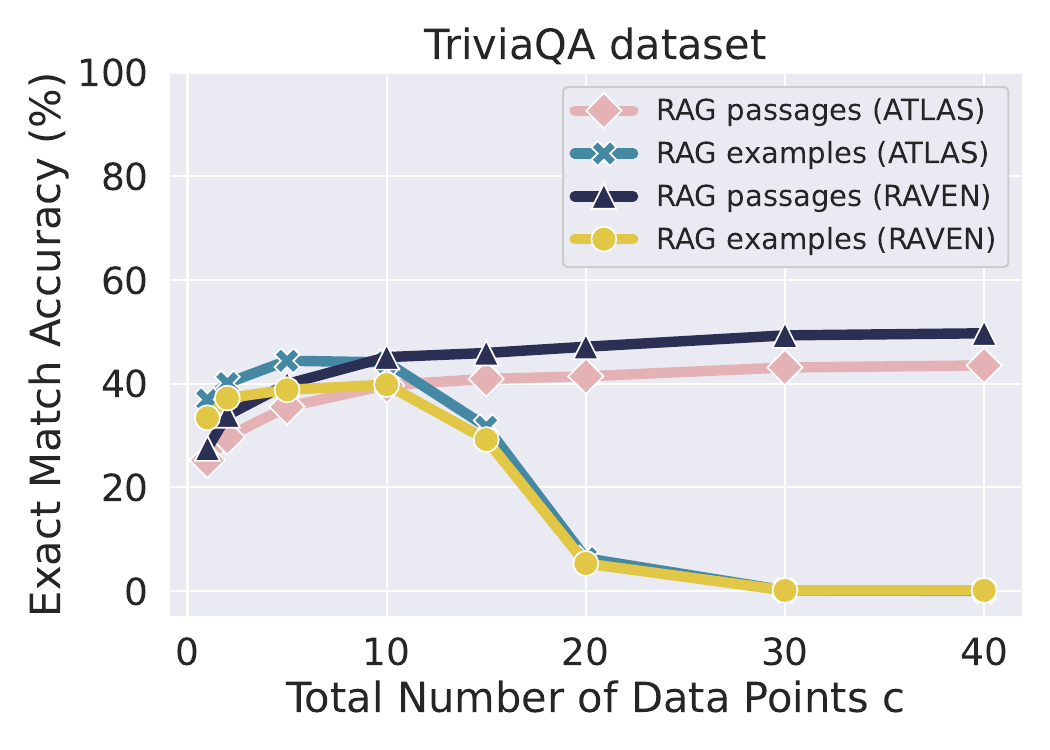}
    \label{fig:exp2_2}
  \end{subfigure}

  \caption{We compare the performance of RAG using examples ($c=n_1$) versus passages ($c=n_2$). }
  \label{fig:rag_ablation2}
\end{figure}

\begin{figure}[h!]
  \centering
  \begin{subfigure}[b]{0.48\textwidth}
    \centering
    \includegraphics[width=\linewidth]{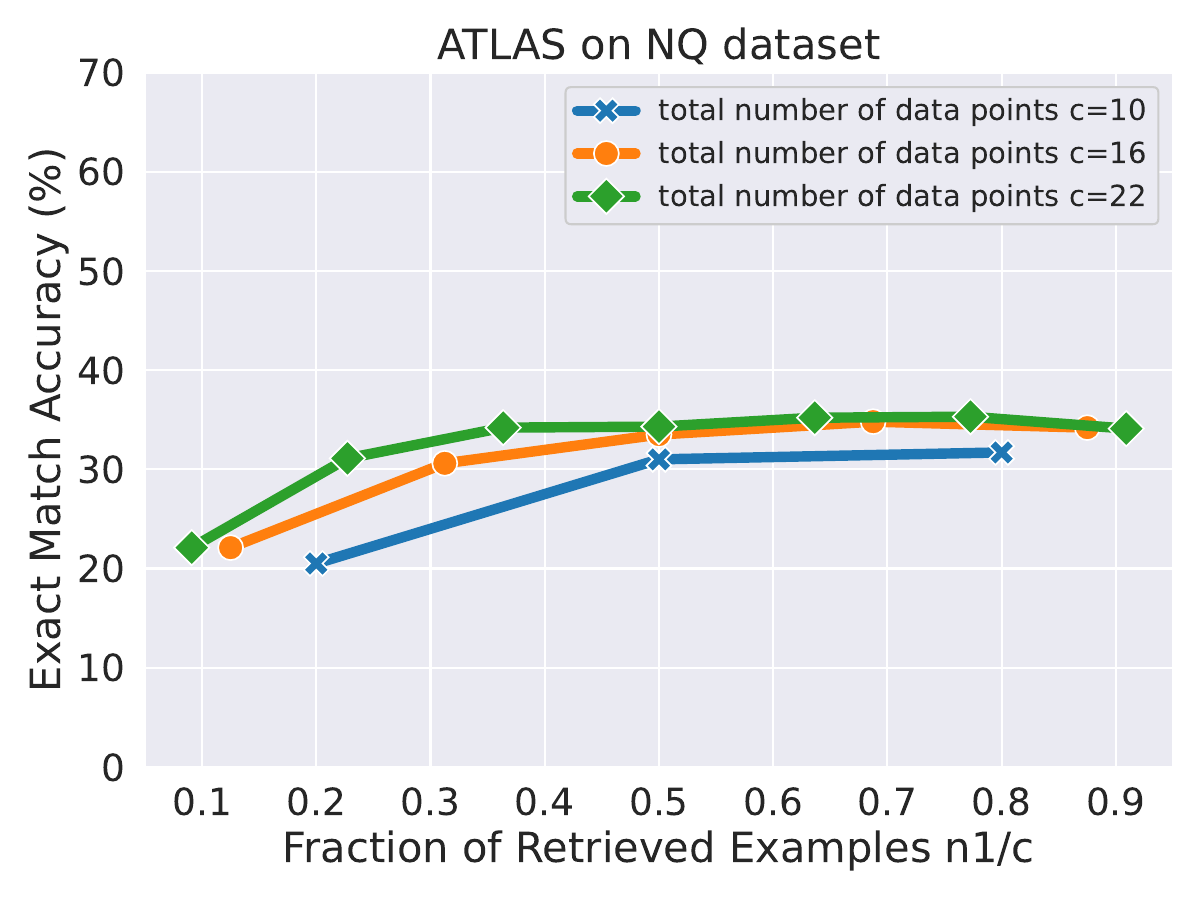}
    \caption{ATLAS Performance as a function of $n_1/c$ under different data points $c$ on NQ.}
    \label{fig:exp3_1}
  \end{subfigure}
  \hfill
  \begin{subfigure}[b]{0.48\textwidth}
    \centering
    \includegraphics[width=\linewidth]{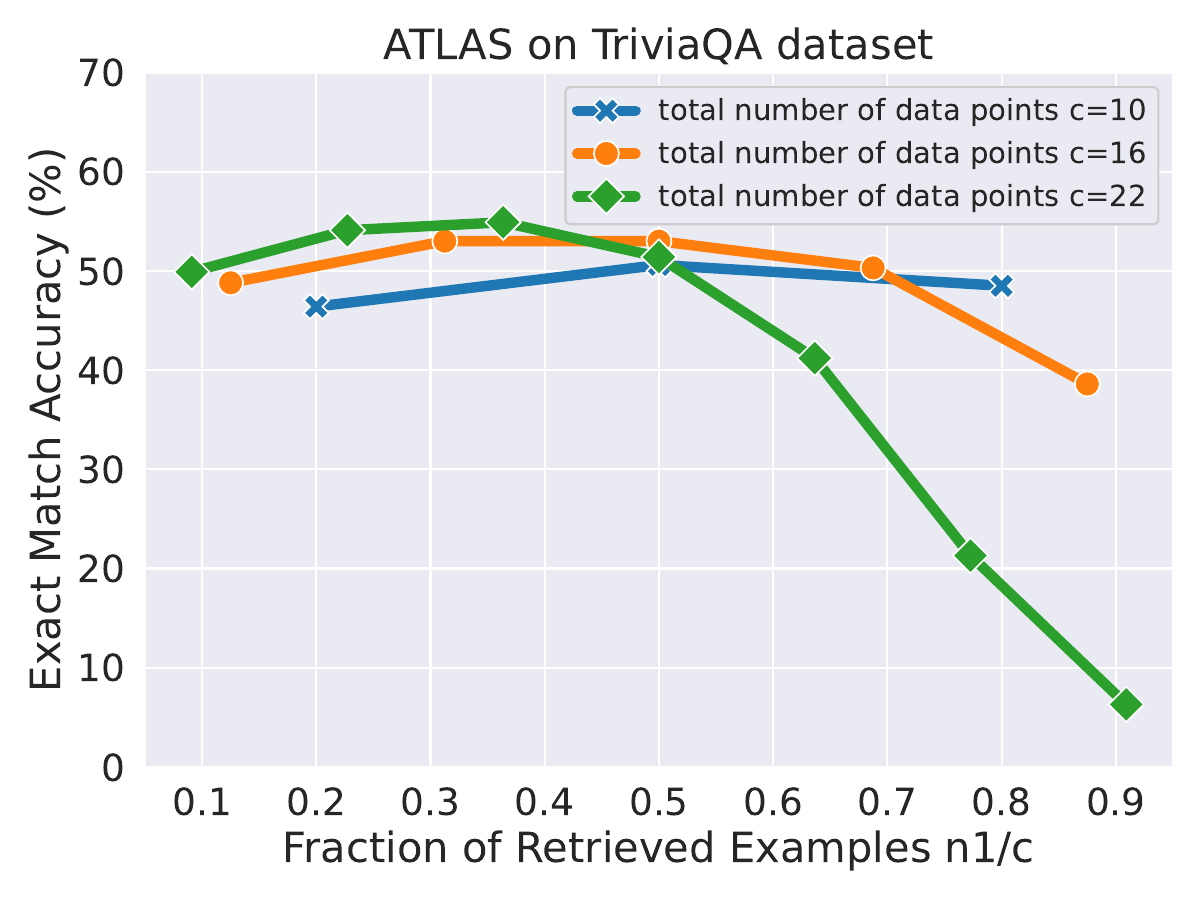}
    \caption{ATLAS Performance as a function of $n_1/c$ under different data points $c$ on TriviaQA.}
    \label{fig:exp3_2}
  \end{subfigure}

  \vspace{1em}

  \begin{subfigure}[b]{0.48\textwidth}
    \centering
    \includegraphics[width=\linewidth]{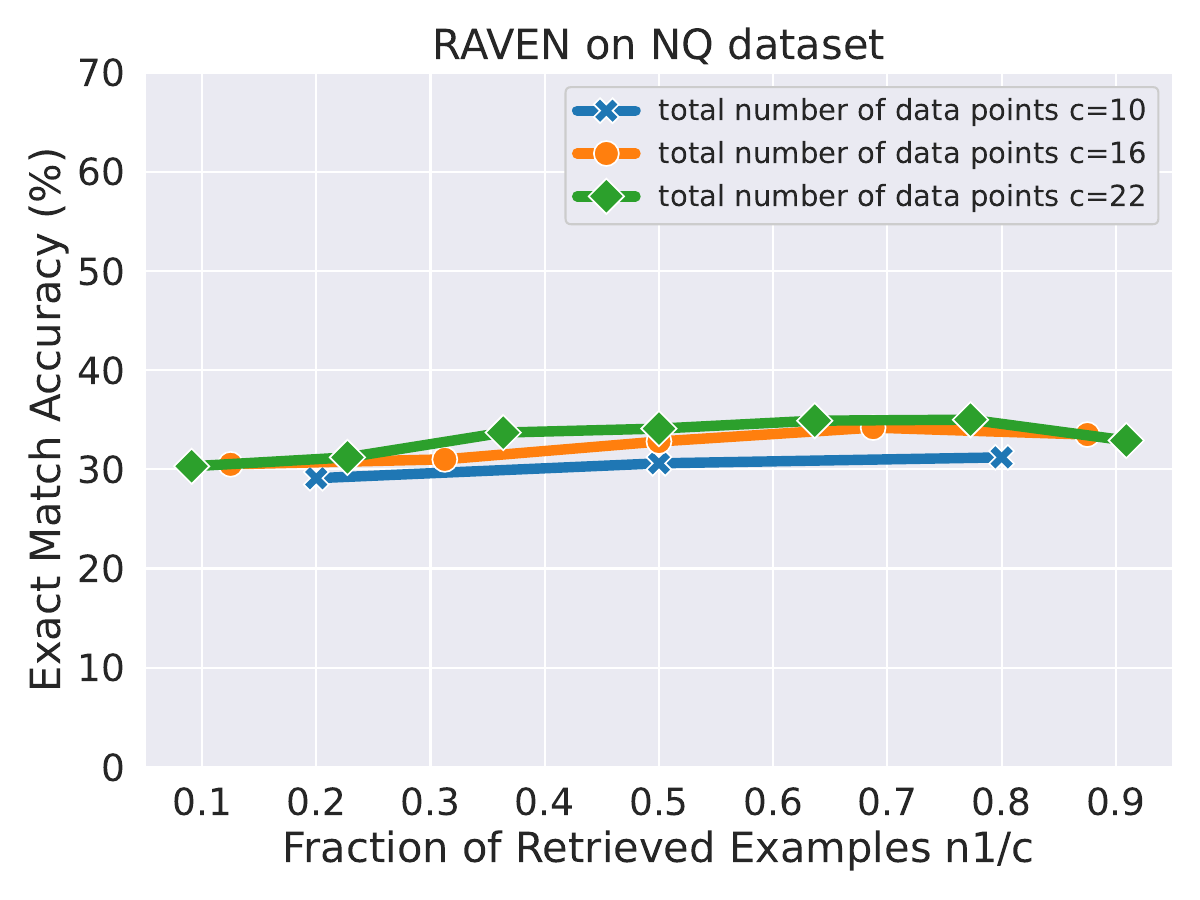}
    \caption{RAVEN Performance as a function of $n_1/c$ under different data points $c$ on NQ.}
    \label{fig:exp3_3}
  \end{subfigure}
  \hfill
  \begin{subfigure}[b]{0.48\textwidth}
    \centering
    \includegraphics[width=\linewidth]{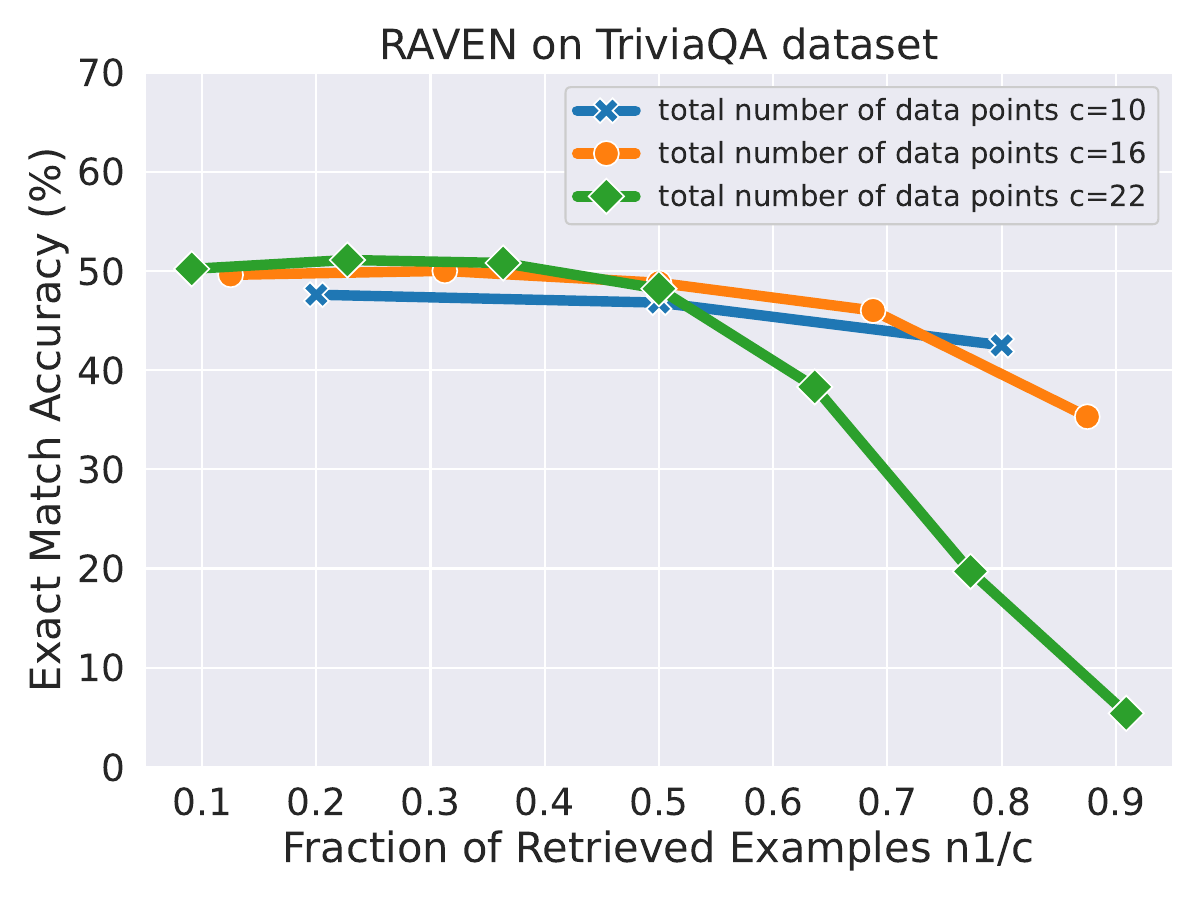}
    \caption{RAVEN Performance as a function of $n_1/c$ under different data points $c$ on TriviaQA.}
    \label{fig:exp3_4}
  \end{subfigure}

  \caption{
    Performance sensitivity to the ratio $n_1/n$ under different data points $c$, where $n_1$ refers to retrieved examples and $n_2$ to passages. 
  }
  \label{fig:exp3}
\end{figure}

\paragraph{RAG Examples v.s.\ RAG External Passages}
Next, we compare using RAG examples from training data only ($m=0, n_1=c, n_2=0$) and RAG passages from external corpora only ($m=0, n_1=0, n_2=c$). The results in Figure~\ref{fig:rag_ablation2} show interesting patterns. For RAG examples only, with more examples, the performance first significantly improves but later drops. This suggests that the first few examples are highly relevant but later ones contain more noise than signal. In contrast, for RAG passages only, the performance increases more slowly but steadily for larger budgets. This suggests the passages retrieved are noisy but still have relevant signals. This aligns with our noise modeling.
When $n_1$ is small ($\leq 20$ for NQ and $\leq 10$ for TriviaQA), RAG examples resemble \emph{uniform noise} due to the relevance of retrieved examples. As $n_1$ increases, $n_1$ introduces more irrelevant or conflicting examples (i.e., \emph{non-uniform noise}). On the other hand, $n_2$ resembles a \emph{uniform noise} regime as the retrieval pool is broad with relevant data but also noisy.

When the retrieval budget is small, retrieval from training examples yield higher accuracy than from passages, even though both operate in the uniform-noise regime. This discrepancy follows from the mixture-noise effects: a passage judged relevant may still lack any answer-bearing text, raising its effective noise level relative to examples.
Furthermore, the significant drop for the retrieval from examples as opposed to retrieval from passages can be explained by the size difference for the training data and passages pool (i.e. Wikipedia). Since the passages provide a denser coverage of the semantic space, more passages will remain relevant as opposed to examples. In all, our theory covers both practical data types and matches the empirical results.

\paragraph{Ratio between RAG Examples and Passages}
The different noise properties of the two kinds of RAG data imply that we should find a proper ratio between them when the total budget $c$ is fixed. Figure~\ref{fig:exp3} in the appendix shows that as the ratio $n_1/c$ increases, the performance initially improves—benefiting from signal information—but eventually declines as low-quality examples dominate the context. This again supports our theoretical view of signal versus noise in the retrieved data. The results demonstrate that performance initially improves as more signal (examples) is added, but eventually declines due to increasing noise from irrelevant or low-quality examples. This supports the theoretical perspective of balancing signal and noise in retrieval-augmented inputs.

\section{Conclusion and Limitations} 

We model RAG as query-dependent noisy in-context learning, derive the first finite-sample error bounds for linear regression that isolate the contributions of retrieval signal and noise, and extend those bounds to different noise regimes and test-time training. Experiments on Natural Questions, TriviaQA with RAVEN, and ATLAS corroborated our theoretical analysis.

Regarding limitations, our bounds focus on the linear setting, opening avenues for future studies on nonlinear methods like kernels and neural networks. While our framework accounts for common RAG noise models, new models may be needed for other types of RAG data. A further direction is to combine RAG with test-time training, studying how on-the-fly adaptation affects both theoretical guarantees and empirical performance. Our experiments feature representative models and datasets, but future research can explore newer retrievers, LLMs like Qwen 3 and Llama 4, and more advanced RAG applications.

\section{Acknowledgment}
This work is partially supported by the National Science Foundation (NSF) under Grant CCF-2046710. I also thank my colleague Haoyue Bai for insightful discussions.

\newpage    
\bibliographystyle{plainnat}
\bibliography{ref}

\newpage
\appendix
\section{Technical Preliminaries}\label{sec:app_prelim}
\paragraph{Additional Notations} 
For two integer indices $i$ and $j$, we denote $\delta_{ij} = \begin{cases}
    1 & \text{   if $i = j$} \\ 
    0 & \text{   if $i\neq j$}
\end{cases}$ as Kronecker delta.

\begin{lemma}[Adapt $\bW$ to Different Context Size]\label{lem:adaptW}
Suppose $\bar{\bW}$ is the weight with context length $m$, then the induced $\bW$ when evaluating on context of length $m'$ is: 
\[
\bW =\frac{m}{m'} \bar{\bW}\]

\end{lemma}

\begin{proof}

We note that $\bar{\bW}$ is the un-normalized weight, i.e. scaling with the inverse context size $1/m$. Only the normalized weight is preserved when applying to a sentence with a different context length.

Then, the prediction is given as: 

\begin{equation*}
\begin{aligned}
\hat{y} &:= \bx_q^\T \bW \bX^\T \by \\ 
&= \bx_q^\T \frac{1}{m'} \bW_{\text{Normalized}} \bX^\T \by \\ 
&= \bx_q^\T \underbrace{\frac{1}{m'} m \bar{\bW}}_{= \bW} \bX^\T \by \\ 
\end{aligned}
\end{equation*}

Thus, 
\[
\bW =\frac{m}{m'} \bar{\bW} \]
\end{proof}

\begin{lemma}[Mixed 4th-Order Moment of Gaussian]\label{lem:mixed-4th-order}
Suppose $\bx \sim \calN(0, I),  \br \sim \calN(0, \delta^2 I)$, then
\begin{enumerate}
    \item \begin{equation}\label{eq:M4_4order_term}
    \begin{aligned}\Exp [\br\br^\T \bW^\T \bx \bx^\T \bW \br\br^\T ] &= 2\delta^4 \bW^\T\bW + \delta^4 \tr(\bW^\T \bW)I 
    \end{aligned}
\end{equation}
    \item \begin{equation}\label{eq:M4_2order_term1}
    \begin{aligned}
    \Exp [\br\bx^\T \bW^\T \bx \bx^\T \bW \bx\br^\T ] = \left(\tr\left(\bW^2 \right) + \tr\left( \bW^\T \bW\right) + \tr\left(\bW \right)^2\right)\delta^2 I \\ 
    \end{aligned}
\end{equation}
\item 
\begin{equation}\label{eq:M4_2order_term2}
\begin{aligned}
\Exp [\bx\br^\T \bW^\T \bx \bx^\T \bW \br\bx^\T ] =  2 \delta^2  \bW\bW^\T+\delta^2  \tr(\bW^\T\bW)I
\end{aligned}
\end{equation}
\item 
\begin{equation}\label{eq:M4_2order_term3}
\Exp [\br\bx^\T \bW^\T \bx \bx^\T \bW \br\bx^\T ] =  \delta^2 \left(\bW^\T\bW + \bW^\T\bW^\T + \bW^\T \tr(\bW) \right)
\end{equation}
\item 
\begin{equation}\label{eq:M4_2order_term4}
\Exp [\br\br^\T \bW^\T \bx \bx^\T \bW \bx\bx^\T ] =  \delta^2 \left(\bW^\T\bW + \bW^\T\bW^\T + \bW^\T \tr(\bW) \right)
\end{equation}
\end{enumerate}
\end{lemma}

\begin{proof}
\begin{enumerate}
    \item We have \begin{equation}
    \begin{aligned}\Exp_{\bx, \br} [\br\br^\T \bW^\T \bx \bx^\T \bW \br\br^\T ] &= \Exp_\br [\br\br^\T \bW^\T  \bW \br\br^\T ] \\ 
    &= 2 \delta^2 I \bW^\T \bW \delta^2 I+ \tr(\bW^\T  \bW \delta^2 I)\delta^2 I \\
    &= 2 \delta^4 \bW^\T  \bW + \delta^4 \tr(\bW^\T  \bW)I  \\ 
    &= 2\delta^4 \bW^\T\bW + \delta^4 \tr(\bW^\T \bW)I 
    \end{aligned}
\end{equation}
where the first step follows from \cref{eq:4th-gaussian1}.
    \item \begin{equation}
    \begin{aligned}
    \Exp_{\bx, \br} [\br\bx^\T \bW^\T \bx \bx^\T \bW \bx\br^\T ] &= \Exp_\br \left[\br \Exp_{\bx} \left[\bx^\T \bW^\T \bx \bx^\T \bW\bx\right] \br^\T \right] \\
    & = \left(\tr\left(\bW^2 \right) + \tr\left( \bW^\T \bW\right) + \tr\left(\bW \right)^2\right)\delta^2 I \\ 
    \end{aligned}
\end{equation}
where the first step follows from \cref{eq:4th-gaussian2}. 
\item 
\begin{equation}
\begin{aligned}
\Exp [\bx\br^\T \bW^\T \bx \bx^\T \bW \br\bx^\T ] 
&=  \Exp \left[\bx\Exp\left[ \tr \left( \br^\T \bW^\T \bx \bx^\T \bW \br \right)\right]\bx^\T \right]  \\ 
&=  \Exp \left[\bx\Exp\left[ \tr \left( \br \br^\T \bW^\T \bx \bx^\T \bW \right)\right]\bx^\T \right] 
\\&= \delta^2 \Exp\bx\tr(\bW^\T \bx \bx^\T \bW ) \bx^\T  
\\&= \delta^2 \Exp\bx\tr(\bx^\T \bW \bW^\T \bx) \bx^\T  
\\&= \delta^2 \Exp \bx \bx^\T \bW\bW^\T \bx\bx^\T 
\\&= 2 \delta^2  \bW\bW^\T+\delta^2  \tr(\bW\bW^\T )I
\\&= 2 \delta^2  \bW\bW^\T+\delta^2  \tr(\bW^\T\bW )I
\end{aligned}
\end{equation}
where the first three steps follow from the cyclic property of trace and the last step follows from \cref{eq:4th-gaussian1}. 
\item 
\begin{equation*}
\begin{aligned}
\Exp[\br\bx^\T \bW \bx \bx^\T \bW \br\bx^\T] &= \Exp\left[\br  (\bx^\T \bW \br)^\T \bx^\T \bW \bx \bx^\T\right] \\ 
&= \Exp\left[\br  \br^\T \bW^\T \bx \bx^\T \bW \bx \bx^\T\right] \\ 
&= \delta^2 \Exp\left[\bW^\T \bx \bx^\T \bW \bx \bx^\T\right] \\ 
&= \delta^2 \bW^\T \left(\bW + \bW^\T +  \tr(\bW)I\right) \\ 
&= \delta^2 \left(\bW^\T\bW + \bW^\T\bW^\T + \bW^\T \tr(\bW) \right)
\end{aligned}
\end{equation*}
where the first step follows from $\bx^\T \bW\br$ being scalar, and the third step follows from \cref{eq:4th-gaussian1}.




\item 
\begin{equation}
\begin{aligned}
\Exp [\br\br^\T \bW^\T \bx \bx^\T \bW \bx\bx^\T ] &= \delta^2 \bW^\T \left(\bW + \bW^\T + \tr\left(\bW \right)\right) \\ 
&=\delta^2 \left(\bW^\T\bW + \bW^\T\bW^\T + \bW^\T \tr(\bW) \right)
\end{aligned}
\end{equation}
It follows from the application of \cref{eq:4th-gaussian1}.
\end{enumerate}
\end{proof}

\begin{lemma}[Expectation of 6th-Order Gaussian Monomial]\label{lem:6order_gaussian}
If $\bx \sim \calN(0, I)$, then
\begin{equation}\label{eq:6order_gaussian}
\begin{aligned}
\Exp [\bx \bx^\T A \bx \bx^\T B \bx \bx^\T] &= A B+A B^\T+A^\T B + A^\T B^\T +B^\T A + B^\T A^\T+B A+B A^\T \\ 
&\quad+\tr(B) A +\tr(B) A^\T+\tr(A) B+\tr(A) B^\T\\ 
&\quad+\tr(A) \tr(B) I+\tr\left(A B^\T\right) I+\tr(A B) I \\ 
\end{aligned}
\end{equation}

\begin{equation}\label{eq:6order_gaussian_W}
\begin{aligned}
\Exp [\bx \bx^\T \bW^\T \bx\bx^\T \bW \bx \bx^\T ] &= \Exp [\bx \bx^\T \bW \bx\bx^\T \bW \bx \bx^\T ] \\ 
&=  2\left(\bW^2 + \bW^\T \bW^\T + \bW^\T \bW+ \bW\bW^\T + \tr\left(\bW\right)\bW + \tr\left(\bW\right) \bW^\T\right) \\ 
& + \tr\left(\bW\right)^2 I + \tr\left(\bW^2\right)I + \tr \left(\bW^\T \bW \right)I \\
\end{aligned}
\end{equation}
\end{lemma}

\begin{proof}
Let $T := \Exp [\bx \bx^\T A \bx \bx^\T B \bx \bx^\T]$. Then, let's consider one scalar entry: 

\begin{equation}
T_{i j}=\Exp\left[\sum_{k, \ell, m, n} x_i x_k A_{k \ell} x_{\ell} x_m B_{m n} x_n x_j\right]=\sum_{k, \ell, m, n} A_{k \ell} B_{m n} \cdot \Exp\left[x_i x_k x_{\ell} x_m x_n x_j\right]
\end{equation}

We now need to compute the 6th-order central moment of standard normal variables. This can be computed using the Isserlis' Theorem \citep{isserlis1918formula}: 

\begin{equation}
\Exp [x_1 \cdots x_s] = \sum_{p \in P_s^2} \prod_{(i,j) \in p} \Exp [x_i x_j]
\end{equation}
where $P_s^2$ stands for all distinct ways of partitioning $\{1, \dots, s\}$ into pairs ${i,j}$ (perfect matching), and the product is over the pairs contained in $p$.

We note that the number of perfect matching for $s$ examples is given as: 
\[\#\text{perfect matching} = \frac{s!}{2^{s/2}(s/2)!}\]
where $2^{s/2}$ is for ignoring the ordering inside pairs and $(s/2)!$ is for ignoring the ordering between pairs.

We note that there are $\frac{6!}{2^3 \cdot 3!} = 15$ distinct partitions for the $6$-th order product of Gaussian random variable. Suppose $(x_a, x_b),(x_c, x_d), (x_e, x_f)$ is a valid pairing, then: 
\[\Exp[x_ax_b]\Exp[x_c x_d]\Exp[x_ex_f] = \begin{cases}
    1 & \text{if } a=b, c=d, e = f \\
    0 & \text{else}
\end{cases} = \delta_{ab}\cdot\delta_{cd}\cdot\delta_{ef}
\]
where $\delta_{ij} := \mathbbm{1}[i = j]$ stands for the Kronecker delta. 

Here, we will discuss the result for all 15 distinct pairings: 

\begin{enumerate}
    \item $(i,k)(l,m)(n,j)$ 
    \[
    \sum_{k,l,m,n} A_{kl}B_{mn} = \sum_{m} A_{im}B_{mj} = A_{i\cdot}B_{\cdot j}=  (AB)_{ij}
    \]
    \item $(i, k)(l, n)(m, j)$
        \[
    \sum_{k,l,m,n} A_{kl}B_{mn} = \sum_{m} A_{im}B_{jm} = A_{i\cdot} B_{j\cdot} = (AB^\T)_{ij}
    \]
    \item $(i, k)(l, j)(m, n)$
    \[
    \sum_{k,l,m,n} A_{kl}B_{mn} = \sum_{m} A_{ij}B_{mm}=  \tr(B)A_{ij}
    \]
    \item $(i, l)(k, m)(n, j)$
    \[
    \sum_{k,l,m,n} A_{kl}B_{mn} = \sum_{m} A_{mi}B_{mj} = A_{\cdot i}B_{\cdot j}=  (A^\T B)_{ij}
    \]
    \item $(i, l)(k, n)(m, j)$
    \[
    \sum_{k,l,m,n} A_{kl}B_{mn} = \sum_{k} A_{ki}B_{jk} = A_{\cdot i}B_{j\cdot}=  (A^\T B^\T)_{i,j}
    \]
    \item $(i, l)(k, j)(m, n)$
    \[
    \sum_{k,l,m,n} A_{kl}B_{mn} = \sum_{m} A_{ji}B_{mm} = (A^\T)_{ij}\tr(B)
    \]
    \item $(i, m)(k, l)(n, j)$
    \[
    \sum_{k,l,m,n} A_{kl}B_{mn} = \sum_{k} A_{kk}B_{ij} = \tr(A)B_{ij} 
    \]
    \item $(i,m)(k,n)(l,j)$
    \[
    \sum_{k,l,m,n} A_{kl}B_{mn} = \sum_{k} A_{kj}B_{ik} = A_{\cdot j}B_{i\cdot}= (BA)_{ij}
    \]
    \item $(i, m)(k, j)(l, n)$
    \[
    \sum_{k,l,m,n} A_{kl}B_{mn} = \sum_{l} A_{jl}B_{il} = A_{j \cdot} B_{i\cdot} = (BA^\T)_{ij}
    \]
    \item $(i, n)(k, l)(m, j)$
    \[
    \sum_{k,l,m,n} A_{kl}B_{mn} = \sum_{k} A_{kk}B_{ji} = \tr(A) (B^\T)_{ij} 
    \]
    \item $(i, n)(k, m)(l, j)$
    \[
    \sum_{k,l,m,n} A_{kl}B_{mn} = \sum_{m} A_{mj}B_{mi} = A_{\cdot j} B_{\cdot i} = (B^\T A)_{ij}
    \]
    \item $(i, n)(k, j)(l, m)$
    \[
    \sum_{k,l,m,n} A_{kl}B_{mn} = \sum_{m} A_{jm}B_{mi} = A_{j \cdot} B_{\cdot i} = (B^\T A^\T)_{ij}
    \]
    \item $(i, j)(k, l)(m, n)$
    \[
    \sum_{k,l,m,n} A_{kl}B_{mn} = \sum_{k,m} A_{kk}B_{mm} = \tr(A)\tr(B)\delta_{ij}
    \]
    \item $(i, j)(k, m)(l, n)$
    \[
    \sum_{k,l,m,n} A_{kl}B_{mn} = \sum_{k, l} A_{kl}B_{kl} = \tr(AB^\T) \delta_{ij}
    \]
    \item $(i, j)(k, n)(l, m)$
    \[
    \sum_{k,l,m,n} A_{kl}B_{mn} = \sum_{m, k} A_{km}B_{mk } = \tr(AB)\delta_{ij}
    \]
\end{enumerate}

Summing up all of these 15 terms together, we obtain Eq.~\eqref{eq:6order_gaussian}. Then, we plug in $A=\bW, B= \bW^\T$, we obtain Eq.~\eqref{eq:6order_gaussian_W}. 
\end{proof}

\begin{lemma}[4th-Order Gaussian Monomial]\label{lem:4th-gaussian} 
Let $\bx,\bx_1,\dots,  \bx_m \sim \calN(0, I)$ and $\bX = [\bx_1^\T; \dots;\bx_m^\T]$. Then, we have
\begin{equation}\label{eq:4th-gaussian1}
\begin{aligned}
\Exp \bx \bx^\T \bW \bx\bx^\T &= \bW + \bW^\T  + \tr(\bW)I  \\ 
&= 2 \bW + \tr(\bW)I \quad \text{if $\bW$ is symmetric}
\end{aligned}
\end{equation}
and 
\begin{equation}\label{eq:4th-gaussian1-mat}
\begin{aligned}
\Exp \bX^\T \bX \bW \bX^\T\bX &= m^2 \bW + m\bW^\T + m\tr(\bW)I \\ 
&= m(m+1)\bW  + m\tr(\bW)I \quad \text{if $\bW$ is symmetric}
\end{aligned}
\end{equation}

\begin{equation}\label{eq:4th-gaussian2}
\begin{aligned}
\Exp \bx^\T A \bx \bx^\T B\bx= \tr\left(A \left(B + B^\T\right)\right) + \tr(A)\tr(B)
\end{aligned}
\end{equation}
If $A= \bW^\T, B= \bW$, then
\begin{equation}
\begin{aligned}
\Exp\bx^\T\bW^\T \bx \bx^\T \bW\bx&=  \Exp\bx^\T\bW \bx \bx^\T \bW\bx= \tr(\bW^\T \bW ) + \tr(\bW^2)+ \tr(\bW)^2
\end{aligned}
\end{equation}

\end{lemma}

\begin{proof}

\cref{eq:4th-gaussian1} follows from section 8.2.4 of \citep{petersen2008matrix} by plugging in mean 0 and variance $I$. 

\begin{equation}
\begin{aligned}
\Exp \bX^\T \bX \bW \bX^\T\bX &= \sum_{i}\bx_i \bx_i^\T \bW \bx_i\bx_i^\T + \sum_{i\neq j} \bx_i \bx_i^\T \bW \bx_j \bx_j^\T \\ 
&= m \left(\bW+\bW^\T + \tr(\bW)I\right) + m(m-1)\bW \\ 
&= m^2 \bW + m\bW^\T + m\tr(\bW)I \\ 
&= m(m+1)\bW + m\tr(\bW)I \\ 
\end{aligned}
\end{equation} 
where the second step follows from plugging in~\cref{eq:4th-gaussian1}.

\cref{eq:4th-gaussian2} follows from section 8.2.4 of \citep{petersen2008matrix} by plugging in mean 0 and variance $I$. 

\end{proof}
\section{Additional Proof for RAG}\label{sec:app_rag_proof}

Here, we provide an overview of the organization of the proof. First, we consider the uniform retrieval noise scenario, and compute the population loss for generic $\bW$ in~\cref{thm:pop-loss-rag}. Then, we plug in the special case $\bW^*$ (isotropic pretrained weight), and provide a closed-form loss in~\cref{prop:rag_pop_loss_isotropic}. Then, we analyze its finite sample complexity in~\cref{prop:rag_loss_nonasymptotic} and the optimal RAG examples in relation to ICL examples in~\cref{prop:optimal_n}.

Later on, we provide an finite sample complexity analysis for non-uniform retrieval noise,~\cref{thm: bound_proportion_noise} for Distance Proportional Noise, and~\cref{thm:bound_probablistic_noise} for Distance-Weighted Mixture Noise. 

\subsection{Uniform Retrieval Noise}

\begin{theorem*}[Restatement of~\cref{thm:pop-loss-rag}]
Under Assumption~\ref{ass:gaussian_offset}, \ref{ass:data}, \ref{ass:uniform_noise}, the population loss of the linear self-attention predictor
$\hat{y}_q=\bx_q^\T\boldsymbol W\bX^\T\by$
satisfies
\begin{equation}\label{eq:rag_loss_tradeoff_app}
\mathcal{L}_{\text{tt+rag}}(\bW)=\underbrace{\Exp\left(\Exp\left(\hat{y}_q\right)-\hat{y}_q\right)^2}_{:=\operatorname{err}_{\text {variance}}(\bW)}+\underbrace{\Exp\left(\Exp\left(\hat{y}_q\right)-\Exp\left(y_q\right)\right)^2}_{:=\operatorname{err}_{\text{bias}}(\bW)}+\underbrace{\sigma^2}_{\text {irreducible noise}}
\end{equation}
Specifically,
\begin{equation}\label{eq:rag_loss_detail_bound}
\begin{aligned}
\err_{\text{variance}}(\bW) &= \left[m\sigma^2 + \left( 1  
+\delta^2\right)n\sigma_{\text{rag}}^2\right]\tr(\bW^\T\bW) + n \sigma_{\text{rag}}^2 \tr(\bW^2) + n\sigma_{\text{rag}}^2 \tr(\bW)^2 \\ 
\err_{\text{bias}}(\bW)&= \betatt^\T \left[I -  (n \delta^2 + 2n + m) (\bW+ \bW^\T) - 2n\tr(\bW)I + M_4\right]\betatt \\ 
&=\betatt^\T \left[I -  (n \delta^2 + 2n + m) (\bW+ \bW^\T) - 2n\tr(\bW)I  \right. \\ 
&\quad +\underbrace{\left[n^2\left(2+\delta^2\right)+n\left(m+\delta^2\right)\right]}_{:= c_1}\left(\bW^2+\left(\bW^2\right)^{\top}\right) +\underbrace{2n(n+\delta^2)}_{:= c_2} \bW \bW^{\top} \\ 
& \quad + \underbrace{\left[m^2+m+m n\left(2+2 \delta^2\right)+n^2\left(2+2 \delta^2+\delta^4\right)+n\left(2 \delta^2+\delta^4\right)\right]}_{:=c_3} \bW^{\top} \bW \\ 
&\quad+\underbrace{\left[n^2\left(2+\delta^2\right)+n\left(m+\delta^2\right)\right]}_{:=c_4, \; c_4 = c_1}\left(\tr(\bW)\left(\bW+\bW^{\top}\right)\right) \\ 
&\left. \quad +\underbrace{\left[n^2+n \delta^2\right]}_{:=c_5}\left(\tr(\bW)^2+\tr\left(\bW^2\right)\right) I +\underbrace{\left[m+n^2+n\left(2 \delta^2+\delta^4\right)\right]}_{:= c_6}\tr\left(\bW^{\top} \bW\right) I  \right] \betatt
\end{aligned} 
\end{equation}

\end{theorem*}

\begin{proof}

For computational convenience, I will define the following quantities for Gram matrix: $\bG_0 = \Xicl^\T\Xicl$, $\bG_i := (\bx_q + \br_i)(\bx_q + \br_i)^\T$, and $\bG := \bG_0 + \sum_{i\in [n]} \bG_i$. 

We write down the error explicitly: 

\begin{equation}\label{eq:pop_loss_breakdown}
\begin{aligned}
y_q - \bx_q^\T\bW \bX^\T\by
&= 
\bx_q^\T \betatt + \epsilon_q -\bx_q^\T \bW \bX^\T \bX \betatt-\bx_q^\T \bW \bX^\T \beps  \\ 
& =\bx_q^\T\left(\betatt-\bW \bG \betatt\right)-\bx_q^\T \bW \bX^\T \beps + \epsilon_q\\
& =\bx_q^\T\left(I-\bW \bG \right) \betatt-\bx_q^\T \bW \bX^\T \beps + \epsilon_q
\end{aligned}
\end{equation}

Therefore, the population loss is equal to:

$$
\begin{aligned}
\calL_{\text{tt+rag}}(\bW) & =\Exp_{(\bx_q, y_q),(\bX, \by),\beps} \left[\left(\bx_q^\T\left(I-\bW \bG \right) \betatt-\bx_q^\T \bW \bX^\T \beps\right)^2\right]+\sigma^2 
\end{aligned}
$$

We note that both $\bepsicl$ and $\bepsrag$ are independent of $\bx_q, \bX$ (including $\br$), and $\Exp [\beps] = 0$.

\[
\Exp_{\beps}\left[-2\left(\bx_q^\T\left(I-\bW \bG\right) \betatt\right)\left(\bx_q^\T \bW \bX^\T \beps\right)\right]=0
\]
And therefore, we have the following loss decomposition:
\begin{equation}\label{eq:rag_loss_decomp}
\calL_{\text{tt+rag}}(\bW) =  \Exp_{\bx_q,\bX, \beps} \left[ \left(\bx_q^\T \bW \bX^\T \beps\right)^2\right]+ \Exp_{\bx_q,\bX } \left[ \left(\bx_q^\T\left(I-\bW \bG \right) \betatt\right)^2\right]+\sigma^2
\end{equation}

Then, we compute the mean of the prediction and the label:
\[
\Exp_{\epsilon_q} y_q  = \Exp_{\epsilon_q} \left(\bx_q^\T \betatt + \epsilon_q \right) = \bx_q^\T \betatt 
\]
\[
\begin{aligned}
\Exp_{\beps}\hat{y}_q &=  \Exp \bx_q^\T \bW \bX^\T\by \\  
&=  \Exp \bx_q^\T \bW \bX^\T(\bX\betatt + \beps)  \\ 
&= \Exp \bx_q^\T \bW \bG \betatt \\  
\end{aligned}
\]
And further, we have 
\begin{equation}\label{eq:bias_variance_terms} 
\begin{aligned}
\Exp_{\epsilon_q}\left(y_q - \Exp_{\epsilon_q}y_q \right)^2 &=  \Exp_{\epsilon_q} \left( \bx_q^\T\betatt + \epsilon_q -\bx_q^\T\betatt   \right)^2 = \Exp_{\epsilon_q}\epsilon_q^2 = \sigma^2 \\ 
\left(\hat{y}_q - \Exp_{\beps} \hat{y}_q\right)^2 &= \left( \bx_q^\T \bW \bX^\T(\bX\betatt + \beps)- \bx_q^\T \bW \bX^\T\bX\betatt\right)^2 = \left( \bx_q^\T \bW \bX^\T\beps\right)^2  \\ 
\left(\Exp_{\epsilon_q}(y_q)- \Exp_{\beps} \hat{y}_q \right)^2 &= \left( \bx_q^\T \betatt - \bx_q^\T \bW \bG\betatt\right)^2 = \left( \bx_q^\T (I-\bW \bG)\betatt \right)^2 
\end{aligned}
\end{equation}
If we plug~\cref{eq:bias_variance_terms} into the loss decomposition~\cref{eq:rag_loss_decomp}, we have 
\begin{equation}
\begin{aligned}
\calL_{\text{tt+rag}}(\bW) &=  \Exp_{\bx_q,\bX, \beps} \left[ \left(\bx_q^\T \bW \bX^\T \beps\right)^2\right]+ \Exp_{\bx_q,\bX } \left[ \left(\bx_q^\T\left(I-\bW \bG \right) \betatt\right)^2\right]+\sigma^2 \\ 
&= \underbrace{\Exp\left(\Exp_{\beps}\left(\hat{y}_q\right)-\hat{y}_q\right)^2}_{:=\operatorname{err}_{\text {variance}}(\bW)}+\underbrace{\Exp\left(\Exp_{\epsilon_q}\left(\hat{y}_q\right)-\Exp_{\beps}\left(y_q\right)\right)^2}_{:=\operatorname{err}_{\text{bias}}(\bW)}+\underbrace{\Exp\left(y_q - \Exp_{\epsilon_q}y_q \right)^2}_{\substack{=\sigma^2 \\ \text{(irreducible noise)}}}
\end{aligned}
\end{equation}
and we can obtain the bias-variance tradeoff as given in~\cref{eq:rag_loss_tradeoff_app}.
\paragraph{Compute $\Exp_{\bx_q,\bX,\br, \beps} \left[ \left(\bx_q^\T \bW \bX^\T \beps\right)^2\right]$}

First, we let
\[
z := \bx_q^\top \bW \bX^\top \beps = \sum_{i=1}^{m+n} \bx_q^\top \bW \bx_i \cdot \epsilon_i
\]

Then,
\[
z^2
= \sum_{i,j=1}^{m+n} (\bx_q^\top \bW \bx_i)(\bx_q^\top \bW \bx_j) \epsilon_i \epsilon_j = \sum_{i,j=1}^{m+n} (\bx_i^\top \bW^\T \bx_q)(\bx_q^\top \bW \bx_j) \epsilon_i \epsilon_j
\]

Taking expectation:
\[
\Exp_{\beps}[z^2] =\sum_{i,j=1}^{m+n} (\bx_i^\top \bW^\T \bx_q)(\bx_q^\top \bW \bx_j) \cdot \Exp[\epsilon_i \epsilon_j]
\]

Because the noise terms are independent and zero-mean, we have:
\[
\Exp[\epsilon_i \epsilon_j] =
\begin{cases}
\sigma^2, & i = j \le m \\
\sigma^2_{\text{rag}}, & i = j > m \\
0, & i \ne j
\end{cases}
\]

So only the diagonal terms survive:
\[
\Exp[z^2] = \sum_{i=1}^m \sigma^2 \cdot \Exp\left[(\bx_q^\top \bW \bx_i)^2\right] + \sum_{i=m+1}^{m+n} \sigma^2_{\text{rag}} \cdot \Exp\left[(\bx_q^\top \bW (\bx_q + \br_{i-m}))^2\right]
\]
\begin{itemize}
    \item \textbf{ICL Term: }Since \(\bx_q, \bx_i \sim \calN(0, I)\) and are independent,
\begin{equation}
\begin{aligned}
\Exp[(\bx_q^\top \bW \bx_i)^2] &= \Exp\left[\bx_i^\T \bW^\T \bx_q \bx^\T_q \bW \bx_i \right] \\ 
&= \Exp \left[\tr\left(\bW^\T \bx_q \bx^\T_q \bW \bx_i\bx_i^\T\right) \right] \\
& = \tr(\bW^\T \bW)
\end{aligned}
\end{equation}
where the first step follows from the cyclic property of trace, the last step follows from the symmetry of $\bW$.

\begin{equation}\label{eq:rag_loss_icl_contribution}
\Rightarrow \quad \text{ICL contribution} = m \cdot \sigma^2 \cdot \tr(\bW^\T \bW)
\end{equation}
    \item \textbf{RAG Term: }

Each row in RAG has the form \(\bx_q + \br_i\), so:
\[
\bx_q^\top \bW (\bx_q + \br_i) = \bx_q^\top \bW \bx_q + \bx_q^\top \bW \br_i
\]

Then, we plug in \cref{eq:4th-gaussian2} into the RAG term:
\begin{equation}\label{eq:pop_loss_variance_rag}
\begin{aligned}
\Exp\left[\left(\bx_q^\top \bW (\bx_q + \br_i)\right)^2\right]
&= \Exp[(\bx_q^\top \bW \bx_q)^2] + \Exp[(\bx_q^\top \bW \br_i)^2] + 2 \Exp[\bx_q^\top \bW \bx_q \cdot \bx_q^\top \bW \br_i] \\
&= \Exp[(\bx_q^\top \bW \bx_q)^2] + \Exp[(\bx_q^\top \bW \br_i)^2] \\
&=\Exp[(\bx_q^\top \bW \bx_q)^2] + \delta^2 \cdot \tr(\bW^\top  \bW) \\
&= \left[\tr(\bW^\T \bW) + \tr(\bW^2) + \tr(\bW)^2 \right] + \delta^2 \cdot \tr(\bW^\T \bW) \\
&= \tr(\bW^\T \bW) + \tr(\bW^2) + \delta^2 \tr(\bW^\T \bW) + \tr(\bW)^2
\end{aligned}
\end{equation}
where the second step follows from $\Exp [\br_i] =  0$, the third step follows from the cyclic property of trace.

\[
\Rightarrow \quad \text{RAG contribution} = n \cdot \sigma^2_{\text{rag}} \cdot \left[(1+\delta^2) \tr(
\bW^\T \bW
)+ \tr(\bW^2) + \tr(\bW)^2\right]
\]

\end{itemize}
Thus, we can combine the two terms above and obtain the following: 
\begin{equation}\label{eq:rag_term}
\begin{aligned}
\Exp \left[ \left( \bx^\T_q  \bW \bX^\T \beps \right)^2 \right] &=  \left[m\sigma^2 + \left( 1  
+\delta^2\right)n\sigma_{\text{rag}}^2\right]\tr(\bW^\T\bW) + n \sigma_{\text{rag}}^2 \tr(\bW^2) + n\sigma_{\text{rag}}^2 \tr(\bW)^2
\end{aligned}
\end{equation}

\paragraph{Compute $\Exp_{\bx_q, \bX} \left[ ( \bx^\T_q\left(I-\bW \bG \right) \betatt )^2\right]$}

First, we can expand the expectation and decompose the inner terms into 4 terms: 

\begin{equation}
\begin{aligned}
\Exp_{\bx_q, \bX}\left[\left(I - \bW \bG \right)^\T \bx_q \bx_q^\T \left(I - \bW\bG\right)\right] &= \Exp_{\bx_q, \bX} \left(I-\bG \bW^\T\right) \bx_q \bx_q^\T(I-\bW \bG) \\ 
&=\underbrace{\Exp\bx_q \bx_q^\T}_{:= M_1}-\underbrace{\Exp\bx_q \bx_q^\T \bW \bG}_{:= M_2}-\underbrace{\Exp\bG \bW^\T \bx_q \bx_q^\T}_{:= M_3}+\underbrace{\Exp\bG \bW^\T \bx_q \bx_q^\T \bW \bG}_{:= M_4}
\end{aligned}
\end{equation}

We denote the four pieces $M_1, M_2, M_3, M_4$ in order.
First, we note that: 
$$
M_1=\Exp\left[\bx_q \bx_q^\T\right]=I
$$

Then, we expand out the terms in $M_2$: 

\begin{equation}
\begin{aligned}
\Exp_{\bx_q, \br} \bx_q \bx_q^\T \bW \bG &= \left(\Exp_{\bx_q, \br} \bx_q \bx_q^\T\right) \bW \bG_0 + \Exp_{\bx_q, \br} \bx_q \bx_q^\T \bW \sum_{i=1}^n (\bx_q + \br_i)(\bx_q + \br_i)^\T  \\ 
&= \bW \bG_0 + \Exp_{\bx_q, \br} \bx_q \bx_q^\T \bW \sum_{i=1}^n (\bx_q + \br_i)(\bx_q + \br_i)^\T  \\ 
&= \bW \bG_0 + \Exp_{\bx_q, \br} \bx_q \bx_q^\T \bW \sum_{i=1}^n (\bx_q \bx_q^\T + \br_i\br_i^\T)  \\ 
&= \bW \bG_0 + \Exp_{\bx_q, \br} \bx_q \bx_q^\T \bW \sum_{i=1}^n (\bx_q \bx_q^\T + \delta^2 I)  \\ 
&= \bW \bG_0 + n(\bW + \bW^\T + \tr(\bW)I) + n\delta^2\bW 
\\ 
&= \bW \bG_0 + n(1+\delta^2)\bW + n\bW^\T + n\tr(\bW)I
\end{aligned}
\end{equation}
where the first step follows from the independence between $\bX$ and $\bx_q$, the second step follows from $\Exp \br_i = 0, \; \forall i \in [n]$, the third step follows from the expectation of $\br_i\br_i^\T = \delta^2 I$, and the last step follows from \cref{eq:4th-gaussian1}. Then, 

\begin{equation}
\begin{aligned}
M_2 &=\bW \bG_0 + n(1+\delta^2)\bW + n\bW^\T + n\tr(\bW)I \\
&= (n \delta^2 + n + m) \bW + n\bW^\T +n\tr(\bW)I 
\end{aligned}
\end{equation}
Similarly, $M_3 = M_2^\T = (n\delta^2 + n + m) \bW^\T + n \bW + n \tr(\bW)I$. Now, we perform similar expansion for $M_4$: 

\begin{equation}
\begin{aligned}
M_4 &= \Exp_{\bx_q,\bX} [\bG \bW^\T \bx_q \bx_q^\T\bW \bG]\\
&= \Exp_{\bx_q,\bX} \left[ \left(\bG_0 + \sum_{i \in [n]} \bG_i \right)\bW^\T \bx_q \bx_q^\T\bW \left(\bG_0 + \sum_{i \in [n]} \bG_i \right)\right]\\
&=  \Exp_{\bx_q, \bX} \left[\bG_0 \bW^\T \bx_q \bx_q^\T\bW \bG_0 + \bG_0 \bW^\T \bx_q \bx_q^\T\bW \sum_{i \in [n]}\bG_i + \sum_{i \in [n]} \bG_i \bW^\T \bx_q \bx_q^\T \bW \bG_0 \right. \\  
& \quad +  \left. \sum_{i \in n} \bG_i \bW^\T \bx_q \bx_q^\T\bW \bG_i + \sum_{i,j \in n, i \neq j} \bG_i \bW^\T \bx_q \bx_q^\T\bW \bG_j \right]\\
&= \Exp_{\bx_q,\bX} \left[\bG_0\bW^\T\bx_q\bx_q^\T \bW\bG_0 + n \underbrace{\bG_0 \bW^\T \bx_q \bx_q^\T\bW \bG_i}_{i \in [n]}+ n \underbrace{\bG_i \bW^\T \bx_q \bx_q^\T\bW \bG_0}_{i \in [n]} \right. \\ 
&\left.\quad  + n \underbrace{\bG_i \bW^\T \bx_q \bx_q^\T\bW \bG_i}_{i \in [n]} + n(n-1) \underbrace{\bG_i \bW^\T \bx_q \bx_q^\T\bW \bG_j}_{i,j \in [n], i \neq j} \right]
\end{aligned}
\end{equation}

First, we can compute that:
\begin{equation}\label{eq:rag_term1}
\begin{aligned}
M_{41}:= \Exp_{\bx_q, \bX} [\bG_0\bW^\T \bx_q\bx_q^\T \bW\bG_0] &= \Exp_{\bX}[\bG_0\bW^\T \bW\bG_0]\\
&= m(m+1) \bW^\T  \bW+ m \tr(\bW^\T  \bW)I 
\end{aligned}
\end{equation}
where the last line follows from \cref{eq:4th-gaussian1-mat} and symmetry of $\bW^\T \bW$. Then, $\forall i\in [n]$, we have: 

\begin{equation}
\begin{aligned}
M_{42}:= \Exp_{\bx_q, \bX}\bG_0 \bW^\T \bx_q\bx_q^\T \bW \bG_i &= \Exp_{\bx_q, \bX} \bG_0 \bW^\T \bx_q\bx_q^\T \bW \left( \bx_q + \br_i\right)\left(\bx_q + \br_i \right)^\T \\
&= \Exp_{\bx_q, \bX} \bG_0 \bW^\T \bx_q\bx_q^\T \bW \left( \bx_q\bx_q^\T + \br_i\br_i^\T\right) \\
&= \Exp_{\bx_q, \bX} \bG_0 \bW^\T \left(\bW+\bW^\T  + \tr(\bW) +  \bW\delta^2 \right)\\
&= m \left(\bW^\T  \bW+\bW^\T \bW^\T  + \tr(\bW ) \bW^\T  + \delta^2  \bW^\T \bW \right) \\
&= m\left((1+\delta^2)\bW^\T \bW +\bW^\T \bW^\T + \tr(\bW)\bW^\T \right) \\
\end{aligned}
\end{equation}
where the first steps follows from $\Exp [\br_i] = 0$, the second step follows from \cref{eq:4th-gaussian1}. 

Moreover, we note that $\forall i \in [n]$: 
\begin{equation}
\begin{aligned}
M_{43}:=\Exp_{\bx_q, \bX, \br_i}\bG_i \bW^\T \bx_q \bx_q^\T\bW \bG_i &= (\bx_q + \br_i)(\bx_q +\br_i)^\T \bW^\T \bx_q \bx_q^\T\bW  (\bx_q + \br_i)(\bx_q +\br_i)^\T \\ 
 &= (\bx_q \bx_q^\top + \br_i\bx_q^\T+\bx_q\br_i^\T + \br_i\br_i^\T) \bW^\T \bx_q \bx_q^\T\bW (\bx_q \bx_q^\top + \br_i\bx_q^\T+\bx_q\br_i^\T + \br_i\br_i^\T) \\
  &= \underbrace{\bx_q \bx_q^\top \bW^\T \bx_q \bx_q^\T\bW\bx_q\bx_q^\top}_{\text{0 order in $\br_i$}} + \underbrace{\br_i\br_i^\T\bW^\T \bx_q \bx_q^\T \bW \br_i\br_i^\T}_{\text{4th-order in $\br_i$}} \\ 
  & + \underbrace{(\br_i\bx_q^\T+\bx_q\br_i^\T)\bW^\T \bx_q \bx_q^\T \bW (\br_i\bx_q^\T+\bx_q\br_i^\T)}_{\text{2nd-order in $\br_i$}} \\ 
  & + \underbrace{\br_i \br_i^\T \bW^\T \bx_q \bx_q^\T \bW \bx_q\bx_q^\T + \bx_q \bx_q^\T \bW^\T \bx_q \bx_q^\T \bW \br_i\br_i^\T}_{\text{2nd-order in $\br_i$}} \\ 
\end{aligned}
\end{equation} 
It worth noting that given Gaussian vector $\br_i$, then its monomial of odd order has 0 expectation according to Isserlis' Theorem ~\citep{isserlis1918formula}. And we can thus obtain the third line by keeping only the even order monomials of $\br_i$. 

By adding up \cref{lem:6order_gaussian} and all the terms above, we obtain that: 
\begin{equation}
\begin{aligned}
&\Exp_{\bx_q, \bX, \br_i}\bG_i \bW^\T \bx_q \bx_q^\T\bW \bG_i \\
&= 
\left.\begin{aligned}
    & 2\left(\bW^2 + (\bW^2)^\T  + \bW^\T \bW+ \bW\bW^\T + \tr\left(\bW\right)(\bW + \bW^\T)\right)  \\ 
    & + \tr\left(\bW\right)^2 I + \tr\left(\bW^2\right)I + \tr \left(\bW^\T \bW \right)I \\ 
\end{aligned} \right\} \text{0th-order in $\br_i$, \cref{lem:6order_gaussian}}\\ 
&\quad+ \underbrace{2\delta^4 \bW^\T\bW + \delta^4 \tr(\bW^\T \bW)I}_{\text{4th-order in $\br_i$, \cref{eq:M4_4order_term}}} \\ 
&\quad+ \underbrace{\delta^2 \left[\tr\left(\bW \right)\left(\bW^\T + \bW \right) +\bW^2+(\bW^2)^\T+ 2\bW^\T \bW  \right]}_{\text{\cref{eq:M4_2order_term4} and its transpose}} \\ 
&\quad+ \underbrace{\left(\tr\left(\bW^2 \right) + \tr\left( \bW^\T \bW\right) + \tr\left(\bW \right)^2\right)\delta^2 I}_{\text{\cref{eq:M4_2order_term1}}} \\ 
&\quad+ \underbrace{ 2 \delta^2  \bW\bW^\T+\delta^2  \tr(\bW^\T\bW )I}_{\text{\cref{eq:M4_2order_term2}}} \\ 
&\quad+ \underbrace{\delta^2 \left[\tr\left(\bW \right)\left(\bW^\T + \bW \right) +\bW^2+(\bW^2)^\T+ 2\bW^\T \bW\right]}_{\text{\cref{eq:M4_2order_term3} and its transpose}} \\
&= (2+2\delta^2) \left[\tr\left(\bW \right)\left(\bW^\T + \bW \right) +\bW^2+(\bW^2)^\T\right]  \\ 
&\quad + (2+4\delta^2)\bW^\T \bW + 2\bW\bW^\T \\ 
&\quad + (1+\delta^2)\left[\tr\left(\bW\right)^2 I + \tr\left(\bW^2\right)I + \tr \left(\bW^\T \bW \right)I \right]\\ 
&\quad+ 2\delta^4 \bW^\T\bW + \delta^4 \tr(\bW^\T \bW)I +  2 \delta^2  \bW\bW^\T+\delta^2  \tr(\bW^\T\bW )I \\
&= (2+2\delta^2) \left[\tr\left(\bW \right)\left(\bW^\T + \bW \right) +\bW^2+(\bW^2)^\T \right]  \\
&\quad +(2+4\delta^2+2\delta^4)\bW^\T\bW + (2+2\delta^2)\bW\bW^\T \\ 
&\quad+(1+\delta^2)\left(\tr\left(\bW\right)^2 + \tr\left(\bW^2\right)\right)I + (1+2\delta^2+\delta^4) \tr(\bW^\T\bW)I \\ 
\end{aligned}
\end{equation}

Also, we expand the cross-term out for $\forall i,j \in [n], i\neq j$: 
\begin{equation}
\begin{aligned}
M_{44}:= \Exp\bG_i \bW^\T \bx_q \bx_q^\T\bW \bG_j &= \Exp (\bx_q + \br_i)(\bx_q +\br_i)^\T \bW^\T \bx_q \bx_q^\T\bW  (\bx_q + \br_j)(\bx_q + \br_j)^\T \\
 &= \left(\bx_q \bx_q^\T +\br_i\br_i^\T\right) \bW^\T \bx_q \bx_q^\T\bW  \left(\bx_q \bx_q^\T + \br_j \br_j^\T \right) \\
 &= \bx_q \bx_q^\T \bW^\T \bx_q \bx_q^\T \bW \bx_q \bx_q^\T + \br_i\br_i^\T \bW^\T \bx_q \bx_q^\T \bW \bx_q\bx_q^\T \\ 
 &\quad + \bx_q\bx_q^\T \bW^\T \bx_q \bx_q^\T \bW \br_j \br_j^\T + \br_i \br_i^\T \bW^\T \bx_q \bx_q^\T \bW \br_j \br_j^\T  \\ 
&= 2\left(\bW^2 +\left(\bW^2\right)^\T+ \bW^\T \bW+ \bW\bW^\T + \tr\left(\bW\right)\bW + \tr\left(\bW\right) \bW^\T\right)  \\ 
& \quad + \tr\left(\bW\right)^2 I + \tr\left(\bW^2\right)I + \tr \left(\bW^\T \bW \right)I\\
& \quad + \delta^2\left(\bW^2 + (\bW^2)^\T+2\bW^\T \bW \right) + \tr(\bW)(\bW + \bW^\T) + \delta^4 \bW^\T \bW\\ 
& = (2 +\delta^2) \left(\bW^2 +\left(\bW^2\right)^\T+ \tr\left(\bW\right)\bW + \tr\left(\bW\right) \bW^\T\right) \\
&\quad + (2+2\delta^2) \bW^\T \bW + 2\bW\bW^\T \\ 
&\quad + \tr\left(\bW\right)^2 I + \tr\left(\bW^2\right)I + \tr \left(\bW^\T \bW \right)I+\delta^4 \bW^\T \bW\\
& = (2 +\delta^2) \left(\bW^2 +\left(\bW^2\right)^\T+ \tr\left(\bW\right)\bW + \tr\left(\bW\right) \bW^\T\right) \\
&\quad + (2+2\delta^2+\delta^4) \bW^\T \bW + 2\bW\bW^\T \\ 
&\quad + \tr\left(\bW\right)^2 I + \tr\left(\bW^2\right)I + \tr \left(\bW^\T \bW \right)I\\
\end{aligned}
\end{equation}
where the first step follows from the independence of $\bx_q, \br_i, \br_j$, and the second step follows from applying \cref{lem:6order_gaussian} and \cref{eq:4th-gaussian1}. 

Combining the above terms together, we have 
\begin{equation}\label{eq:M4}
\begin{aligned}
M_4 &= M_{41} + n(M_{42} + M_{42}^\T) + nM_{43} + n(n-1)M_{44} \\ 
&= m(m+1) \bW^\T  \bW  + m \tr(\bW^\T  \bW)I + mn\left(\left(2+ 2 \delta^2\right)\bW^\T \bW+\bW^2 + (\bW^2)^\T  + \tr\left(\bW\right)\left(\bW + \bW^\T\right)\right)\\ 
&\quad + nM_{43} + n(n-1)M_{44} \\ 
&= 2n\left(2n+\delta^{2}\right)\bW^{2} +2n\left(n+\delta^{2}\right)\bW\bW^\T
\\
&\quad+\left[
      m^{2}+m
      + (4+2\delta^{2})m n
      + n^{2}\left(2+4\delta^{2}+\delta^{4}\right)
      + n\left(2\delta^{2}+\delta^{4}\right)
  \right]\bW^\T\bW
\\
&\quad+\left[
      n^{2}\left(2+\delta^{2}\right)
      + n\left(m+\delta^{2}\right)
  \right]
 \tr(\bW)\left(\bW+\bW^\T\right)
\\
&\quad+\left(n^{2}+n\delta^{2}\right)
  \left(\tr(\bW)^{2}+\tr(\bW^{2})\right)I+\left[m + n^{2} +n\left(2\delta^{2}+\delta^{4}\right)\right]\tr\left(\bW^\T\bW\right)I \\
& = \left[n^2\left(2+\delta^2\right)+n\left(m+\delta^2\right)\right]\left(\bW^2+\left(\bW^2\right)^{\top}+\tr(\bW)\left(\bW+\bW^{\top}\right)\right) \\ 
&\quad +\left[2 n^2+2 n \delta^2\right] \bW \bW^{\top} \\ 
& \quad + {\left[m^2+m+m n\left(2+2 \delta^2\right)+n\left(2 \delta^2+\delta^4\right)+n^2\left(2+2 \delta^2+\delta^4\right)\right] \bW^{\top} \bW} \\ 
&\quad +\left[n^2+n \delta^2\right]\left(\tr(\bW)^2+\tr\left(\bW^2\right)\right) I \\ 
& \quad +\left[m+n^2+n\left(2 \delta^2+\delta^4\right)\right] \tr\left(\bW^{\top} \bW\right) I 
\end{aligned}
\end{equation}

In summary, combining all terms together, we have: 

\[
\calL(\bW) := \err_{\text{variance}} +\err_{\text{bias}} + \sigma^2
\]
where the \textit{irreducible variance} is $\sigma^2$, and the \textit{reducible variance} (variance of ICL + RAG) is 
\[
\text{Variance of ICL} + \text{Variance of RAG} = 
\left[m\sigma^2 + \left( 1  
+\delta^2\right)n\sigma_{\text{rag}}^2\right]\tr(\bW^\T\bW) + n \sigma_{\text{rag}}^2 \tr(\bW^2) + n\sigma_{\text{rag}}^2 \tr(\bW)^2
\]

And the err from the bias term is given as: 

\[
\begin{aligned}
\err_{\text{bias}} &=\betatt^\T [M_1 - M_2 - M_3 + M_4] \betatt\\ 
&= \betatt^\T \left[I -  (n \delta^2 + 2n + m) (\bW+ \bW^\T) - 2n\tr(\bW)I + M_4\right]\betatt \\ 
&=\betatt^\T \left[I -  (n \delta^2 + 2n + m) (\bW+ \bW^\T) - 2n\tr(\bW)I  \right. \\ 
&\quad +\left[n^2\left(2+\delta^2\right)+n\left(m+\delta^2\right)\right]\left(\bW^2+\left(\bW^2\right)^{\top}\right) +2n(n+\delta^2) \bW \bW^{\top} \\ 
& \quad + {\left[m^2+m+m n\left(2+2 \delta^2\right)+n^2\left(2+2 \delta^2+\delta^4\right)+n\left(2 \delta^2+\delta^4\right)\right] \bW^{\top} \bW} \\ 
&\quad+\left[n^2\left(2+\delta^2\right)+n\left(m+\delta^2\right)\right]\left(\tr(\bW)\left(\bW+\bW^{\top}\right)\right) \\ 
&\left. \quad +\left[n^2+n \delta^2\right]\left(\tr(\bW)^2+\tr\left(\bW^2\right)\right) I +\left[m+n^2+n\left(2 \delta^2+\delta^4\right)\right] \tr\left(\bW^{\top} \bW\right) I  \right] \betatt
\end{aligned}
\]
\end{proof}
The previous theorem gives the exact form the RAG population with general $\bW$. In the following proposition, we will compute the population under special $\bW$ in order to obtain a more fine-grained complexity analysis.

\begin{proposition}[RAG Population loss under isotropic setting] \label{prop:rag_pop_loss_isotropic}
Assuming $\bW^* = \frac{m}{(m+d+1)(m+n)} I$. Then, the population loss are given as: 
\begin{equation*}
\begin{aligned}
\calL_{\text{tt+rag}}(\bW^*) &= \err_{\text{variance}}(\bW^*) + \err_{\text{bias}}(\bW^*) + \sigma^2 \\ 
\err_{\text{variance}}(\bW^*) &= \frac{m^3 d}{[(m+d+1)(m+n)]^2} \sigma^2+\frac{dm^2 n (2+\delta^2+d)}{[(m+d+1)(m+n)]^2} \sigma_{\mathrm{rag}}^2 \\ 
\err_{\text {bias}}(\bW^*)&=\left\|\beta_{T T}\right\|_2^2\left[1-\frac{2 m}{(m+d+1)(m+n)}\left(n \delta^2+2 n+m+n d\right)+\frac{P(m,n,d,\delta)m^2}{(m+d+1)^2(m+n)^2}\right] 
\end{aligned}
\end{equation*}
where 
\[
\begin{aligned}
P (m,n,d,\delta)= & 6 n^2+4 n \delta^2+m^2+m+\left(4+2 \delta^2\right) m n \\
& \quad+n^2\left(2+4 \delta^2+\delta^4\right)+n\left(2 \delta^2+\delta^4\right) +2 d n^2\left(2+\delta^2\right)+2 d n\left(m+\delta^2\right) \\
& \quad +d(d+1)\left(n^2+n \delta^2\right)+d m+d n^2+d n\left(2 \delta^2+\delta^4\right)
\end{aligned}
\]
\end{proposition}
\begin{proof}
Plugging in the value of $\bW^*$, we first compute the error from input variance.
\begin{equation*}
\begin{aligned}
\tr((\bW^*)^2) &= \frac{dm^2}{(m+d+1)^2(m+n)^2} \\ 
\tr(\bW^*) &= \frac{dm}{(m+d+1)(m+n)}
\end{aligned}
\end{equation*} 

\[
\begin{aligned}
\err_{\text{variance}}(\bW^*)&= \left[m \sigma^2+\left(1+\delta^2\right) n \sigma_{r a g}^2\right] \tr\left(\bW^{\top} \bW\right)+n \sigma_{r a g}^2 \tr\left(\bW^2\right)+n \sigma_{r a g}^2 \tr(\bW)^2 \\ 
&= \frac{dm^2[m\sigma^2 + (1+\delta^2)n\sigma_{\text{rag}}^2]}{(m+d+1)^2(m+n)^2} + n\sigma_{\text{rag}}^2 \frac{dm^2}{(m+d+1)^2(m+n)^2}  + n \sigma_{\text{rag}}^2 \frac{d^2m^2}{(m+d+1)^2(m+n)^2} \\
&= \frac{m^3 d}{[(m+d+1)(m+n)]^2} \sigma^2+\frac{dm^2 n (2+\delta^2+d)}{[(m+d+1)(m+n)]^2} \sigma_{\mathrm{rag}}^2
\end{aligned}
\]

Then, we proceed to plug in the value and compute the error from the estimation bias.

\[
\begin{aligned}
\err_{\text {bias}}(\bW^*)&= 
\|\betatt\|_2^2 \left[ 1-\frac{2m(n\delta^2+ 2n + m)}{(m+n)(m+d+1)} -\frac{2ndm}{(m+n)(m+d+1)} + \frac{m^2}{(m+d+1)^2(m+n)^2}\underbrace{(\dots)}_{P(m,n,d,\delta)} \right] \\ 
&= \left\|\beta_{T T}\right\|_2^2\left[1-\frac{2 m}{(m+d+1)(m+n)}\left(n \delta^2+2 n+m+n d\right)+\frac{P(m,n,d,\delta)m^2}{(m+d+1)^2(m+n)^2}\right]
\end{aligned}
\]
where
\[
\begin{aligned}
P (m,n,d,\delta) &= (2c_1+c_2+c_3) + 2dc_4 + (d^2 + d)c_5 + dc_6  \\ 
&= 2\left(n^2\left(2+\delta^2\right)+n\left(m+\delta^2\right)\right) +2n(n+\delta^2) \\ 
&\quad+\left[m^2+m+m n\left(2+2 \delta^2\right)+n^2\left(2+2 \delta^2+\delta^4\right)+n\left(2 \delta^2+\delta^4\right)\right]   \\ 
&\quad+2d[n^2(2+\delta^2)+n(m+\delta^2)] + (d^2+d)(n^2+n\delta^2) + d[m+n^2+n(2\delta^2+\delta^4)]\\ 
&= 6 n^2+4 n \delta^2+m^2+m+\left(4+2 \delta^2\right) m n \\
& \quad+n^2\left(2+4 \delta^2+\delta^4\right)+n\left(2 \delta^2+\delta^4\right) +2 d n^2\left(2+\delta^2\right)+2 d n\left(m+\delta^2\right) \\
& \quad +d(d+1)\left(n^2+n \delta^2\right)+d m+d n^2+d n\left(2 \delta^2+\delta^4\right)
\end{aligned}
\]
\end{proof}
\subsubsection{Finite Sample Complexity of RAG} 
\begin{proposition*}[Restatement of~\cref{prop:rag_loss_nonasymptotic}] Under Assumption~\ref{ass:gaussian_offset}, \ref{ass:data}, \ref{ass:uniform_noise}, if $\delta^2 \ll 1$, 
\[
\calL_{\text{tt+rag}}(\bW^*) =\calO \left(\sigma^2 + \underbrace{\frac{dm}{(m+n)^2}\sigma^2 + \frac{d^2n}{(m+n)^2}\sigma_{\text{rag}}^2}_{\err_\var(\bW^*)} + \underbrace{\|\betatt\|_2^2 \left[ \frac{d}{m} + d^2\left(\frac{n}{m+n}\right)^2 \right]}_{\err_\bias(\bW^*)}\right)
\]
\begin{equation}
\begin{aligned}
\err_{\text{variance}}(\bW^*)
&= \begin{cases}
 \calO(\frac{d}{m}\sigma^2 + \frac{d^2}{m^2}\sigma_{\text{rag}}^2 ) = \calO\left(\frac{1}{m}\right) & \text{ $m \rightarrow \infty$, $n$ fixed.}
 \\ 
\calO(\frac{d}{n^2}\sigma^2 + \frac{d^2}{n}\sigma_{\text{rag}}^2) =  \calO\left(\frac{1}{n}\right)& \text{ $n \rightarrow \infty$, $m$ fixed} \\ 
\calO(\frac{d}{m}\sigma^2 + \frac{d^2}{m}\sigma_{\text{rag}}^2) = \calO \left(\frac{1}{m}\right) & \text{ $m,n \rightarrow \infty$, $n = \Theta(m)$}    
\end{cases}
\end{aligned}
\end{equation}

\begin{equation}
\begin{aligned}
\err_{\text{bias}}(\bW^*)
&= \begin{cases}
\calO\left(\|\betatt\|_2^2 \frac{d}{m}  \right) & \text{if $m\rightarrow \infty$, $n$ is fixed} \\ 
 \calO\left(\|\betatt\|_2^2  d^2 \right)  = C_1 &\text{if $n\rightarrow \infty$, $m$ is fixed} \\ 
\calO\left(\|\betatt\|_2^2 \left( \frac{d}{m}+ d^2\right) \right) = C_2 + \calO(\|\betatt\|_2^2\frac{d}{m}) &\text{if $m\rightarrow \infty$, $n = \Theta(m)$} \\ 
\end{cases}
\end{aligned}
\end{equation}
\end{proposition*}

\begin{proof}
We will bound the variance-induced error and the bias-induced error separately.
\paragraph{Variance-Induced Error}
First, we try to bound $\err_{\text{variance}}(\bW^*)$: 
\begin{equation}
\begin{aligned}
\err_{\text{variance}}(\bW^*) &= \frac{dm^3}{(m+d+1)^2(m+n)^2}\sigma^2 + \frac{dm^2n(2+\delta^2+d)}{(m+d+1)^2(m+n)^2}\sigma_{\text{rag}}^2 \\ 
&\leq \frac{dm^3}{m^2(m+n)^2}\sigma^2 + \frac{dm^2n(d+\delta^2+2)}{m^2(m+n)^2}\sigma^2_{\text{rag}} \\ 
&= \frac{dm}{(m+n)^2}\sigma^2 + \frac{d(2+\delta^2+d)n}{(m+n)^2}\sigma_{\text{rag}}^2 \\ 
&= \calO\left(\frac{dm}{(m+n)^2}\sigma^2 + \frac{d^2n}{(m+n)^2}\sigma_{\text{rag}}^2\right) \\ 
&= \begin{cases}
 \calO(\frac{d}{m}\sigma^2 + \frac{d^2}{m^2}\sigma_{\text{rag}}^2 ) = \calO\left(\frac{1}{m}\right) & \text{ $m \rightarrow \infty$, $n$ fixed.}
 \\ 
\calO(\frac{d}{n^2}\sigma^2 + \frac{d^2}{n}\sigma_{\text{rag}}^2) =  \calO\left(\frac{1}{n}\right)& \text{ $n \rightarrow \infty$, $m$ fixed} \\ 
\calO(\frac{d}{m}\sigma^2 + \frac{d^2}{m}\sigma_{\text{rag}}^2) = \calO\left( \frac{1}{m}\right) & \text{ $m,n \rightarrow \infty$, $n = \Theta(m)$}    
\end{cases}
\end{aligned}
\end{equation}
where the second line follows from $(m+d+1)\geq m$ and the fourth line follows from the fact that $\delta^2$ is small relative to $m,n,d$. 

\paragraph{Bias-Induced Error}
We will expand out the term 
\begin{equation}
\begin{aligned}
\err_{\text{bias}}(\bW^*) &= \|\betatt\|_2^2\frac{Q(m,n;d,\delta^2)}{(m+d+1)^2(m+n)^2} \\ 
\end{aligned}
\end{equation}
where 
\begin{equation}\label{eq:rag_Q}
\begin{aligned}
Q(m,n;d,\delta^2) &:= (m+n)^2(m+d+1)^2 - 2m(m+n)(m+d+1)(n\delta^2+2n+m+nd) + m^2 P(m,n,d,\delta^2) \\
& =(d+1)m^3 + \underbrace{(d^2 + 2d\delta^2 + 4d+\delta^4+2\delta^2+5)}_{:= \kappa_{22}}m^2n^2  \\ 
&\quad+ \underbrace{(d^2 \delta^2-2 d^2+d \delta^4+3 d \delta^2-4 d+\delta^4+4 \delta^2-2)}_{:=\kappa_{21}}m^2n \\ 
&\quad\underbrace{-\left(2 d^2+2 d \delta^2+4 d+2 \delta^2+2\right)}_{:=\kappa_{12}}mn^2 + (d^2+2d+1)(m+n)^2 \\ 
& = (d+1)m^3 + \kappa_{22}m^2n^2 + |\kappa_{21}|m^2n + \text{lower-order terms}\\ 
& \leq (d+1)m^3 + \kappa_{22}m^2n^2 + |\kappa_{21}|m^2n +(d+1)^2(m+n)^2\\ 
\end{aligned}
\end{equation}
where the last line follows from $\kappa_{12} <0$.

Note that we assume $\delta^2 \ll 1$. Now, we can bound each of the term in $Q$ divided individually: 
\begin{itemize}
    \item Cubic term: 
    \begin{equation}\label{eq:rag_plugin_bias2}
        \frac{(d+1) m^3}{m^2(m+n)^2}=\frac{d+1}{m}\left(\frac{m}{m+n}\right)^2 \leq \frac{d+1}{m}
    \end{equation}
    \item Skew-cubic term:
    \begin{equation}\label{eq:rag_plugin_bias3}
        \frac{\left|\kappa_{21}\right| m^2 n}{m^2(m+n)^2}=\left|\kappa_{21}\right| \frac{n}{(m+n)^2} \leq\left|\kappa_{21}\right| \frac{n}{(m+n)^2}
    \end{equation}
    \item Quartic term: 
    \begin{equation}\label{eq:rag_plugin_bias4}
        \frac{\kappa_{22} m^2 n^2}{m^2(m+n)^2}=\kappa_{22} \left(\frac{n}{m+n}\right)^2
    \end{equation}
    \item last term: 
    \[
    (d+1)^2(m+n)^2 \frac{1}{m^2(m+n)^2} = \frac{d^2}{m^2}
    \]
\end{itemize}

Combining \cref{eq:rag_Q}, \cref{eq:rag_plugin_bias2}, \cref{eq:rag_plugin_bias3}, \cref{eq:rag_plugin_bias4}, we can obtain that 
\begin{equation}
\begin{aligned}
\err_{\text{bias}}(\bW^*) 
&= \calO\left(\|\betatt\|_2^2 \left[ \frac{dm}{(m+n)^2} + d^2\frac{n^2}{(m+n)^2} + \frac{d^2}{m^2}   \right] \right) \\ 
&= \begin{cases}
\calO\left(\|\betatt\|_2^2 \frac{d}{m}  \right) & \text{if $m\rightarrow \infty$, $n$ is fixed} \\ 
 \calO\left(\|\betatt\|_2^2  d^2 \right)  = C_1 &\text{if $n\rightarrow \infty$, $m$ is fixed} \\ 
\calO\left(\|\betatt\|_2^2 \left( \frac{d}{m}+ d^2\right) \right) = C_2 + \calO(\|\betatt\|_2^2\frac{d}{m}) &\text{if $m\rightarrow \infty$, $n = \Theta(m)$} \\ 
\end{cases}
\end{aligned}
\end{equation}
where the third step follows from plugging in the highest order monomial of $d$ from $\kappa_{21},\kappa_{22}$.

\end{proof}

\subsubsection{Optimality of Number of RAG Examples}
\begin{proposition*}[Restatement of~\cref{prop:optimal_n}]
Under Under Assumption~\ref{ass:gaussian_offset},\ref{ass:data},\ref{ass:uniform_noise}, $\delta^2 \ll 1$, and reasonable choice of $\sigma^2, \sigma_{\text{rag}}^2$ ($\sigma^2, \sigma_{\text{rag}}^2 \ll \|\betatt\|_2^2$), the optimal $n^*$ that minimizes the RAG loss follows: 
\begin{equation}
n^* = \calO\left(\frac{m \left(d^2 \|\betatt\|_2^2 + d \sigma^2- d^2 \sigma_{\text{rag}}^2\right)}{m d^2 \|\betatt\|_2^2 - d^2\sigma_{\text{rag}}^2}\right) = \calO\left(\frac{d\|\betatt\|_2^2 + \sigma^2-d\sigma_\rag^2 }{d\|\betatt\|_2^2}\right)
\end{equation}
and the improvement on loss from picking the optimal $n^*$ over $n =0$ is given as:
\begin{equation}
\calL_{\text{tt+rag}}(\bW^*)|_{n=0} - \calL_{\text{tt+rag}}(\bW^*)|_{n=n^*} = \calO\left(\frac{1}{m^2}\right)
\end{equation}
\end{proposition*}

\begin{proof}

First, we define several constants that can lead to a cleaner calculation. Let $\omega_1:=d$, $\omega_2 := d^2$. Then,
\[
\begin{aligned}
\err_{\text{variance}}(\bW^*) &= \frac{dm^3}{(m+d+1)^2(m+n)^2}\sigma^2 + \frac{dm^2n(2+\delta^2+d)}{(m+d+1)^2(m+n)^2}\sigma_{\text{rag}}^2 \\ 
&\approx \frac{m^3}{(m+d+1)^2(m+n)^2} \omega_1\sigma^2 + \frac{m^2n}{(m+d+1)^2(m+n)^2}\omega_2\sigma_{\text{rag}}^2 \\
\end{aligned}
\]
where the last line follows from $\delta^2 \ll 1$. Let $Q(m,n,d,\delta^2) := \frac{\err_{\text {bias}}(\bW^*)(m+d+1)^2(m+n)^2}{\|\betatt\|_2^2}$ as in~\cref{eq:rag_Q}. Then,
\begin{equation}
\begin{aligned}
Q(m,n;d,\delta^2) &= (m+n)^2(m+d+1)^2 - 2m(m+n)(m+d+1)(n\delta^2+2n+m+nd) + m^2 P(m,n,d,\delta^2) \\
& =(d+1)m^3 + (d^2 + 2d\delta^2 + 4d+\delta^4+2\delta^2+5)m^2n^2  \\ 
&\quad+ (d^2 \delta^2-2 d^2+d \delta^4+3 d \delta^2-4 d+\delta^4+4 \delta^2-2)m^2n \\ 
&\quad-\left(2 d^2+2 d \delta^2+4 d+2 \delta^2+2\right)mn^2 + (d^2+2d+1)(m^2+n^2) \\ 
&\approx \underbrace{d}_{:= \tau_{30}}m^3 + \underbrace{d^2}_{\tau_{22}}m^2n^2 \underbrace{- 2d^2}_{:= \tau_{21}} m^2 n \underbrace{- 2d^2}_{:= \tau_{12}} mn^2 + \underbrace{d^2}_{:=\tau_2}(m^2+n^2) \\
&= \tau_{30}m^3 + \tau_{22}m^2 n^2 + \tau_{21}m^2n + \tau_{12}mn^2 + \tau_2(m^2 + n^2)
\end{aligned}
\end{equation} 

Now, we want to find the optimal $n^*$ w.r.t. $\calL_{\text{tt+rag}}$. That is, we want to minimize
\begin{equation}\label{eq:loss_approx}
\left[m^3\omega_1 \sigma^2 + m^2n \omega_2 \sigma_{\text{rag}}^2+ \|\betatt\|_2^2\left(\tau_{30} m^3+\tau_{22} m^2 n^2+\tau_{21} m^2 n+\tau_{12} m n^2+\tau_2\left(m^2+n^2\right) \right)\right]\frac{1}{(m+n)^2(m+d+1)^2}
\end{equation}
where all $\tau, \omega$ are positive except that $\tau_{12}$ is negative. First, we take out the terms that does not depend on $n$, and we equivalently minimize
\[
L(n) := \left[m^3\omega_1 \sigma^2 + m^2n \omega_2 \sigma_{\text{rag}}^2+ \|\betatt\|_2^2\left(\tau_{30} m^3+\tau_{22} m^2 n^2+\tau_{21} m^2 n+\tau_{12} m n^2+\tau_2\left(m^2+n^2\right) \right)\right]\frac{1}{(m+n)^2}
\]
Let
\begin{equation}
\begin{aligned}
& A=m^3 \omega_1 \sigma^2+\|\betatt\|^2 \tau_{30} m^3+\|\betatt\|^2 \tau_2 m^2, \\
& B=m^2\left(\omega_2 \sigma_{\mathrm{rag}}^2+\|\betatt\|^2 \tau_{21}\right), \\
& C=\|\betatt\|^2\left(\tau_{22} m^2+\tau_{12} m+\tau_2\right) .
\end{aligned}
\end{equation} 
Then, 
\[L(n) = (A + Bn +Cn^2)/(m+n)^2\]
Then, by the rule for derivative of quotient, 
\[
\begin{aligned}
\frac{\partial\left(\calL_{\text{tt+rag}}(\bW^*)\right)}{\partial n} &= \frac{(B+2 C n)(m+n)^2-2(m+n)\left(A+B n+C n^2\right)}{(m+n)^4}\\
& =\frac{(B+2 C n)(m+n)-2\left(A+B n+C n^2\right)}{(m+n)^3} \\
& =\frac{B m+B n+2 C m n+2 C n^2-2 A-2 B n-2 C n^2}{(m+n)^3} \\
& =\frac{B m-B n+2 C m n-2 A}{(m+n)^3} 
\end{aligned}
\]
Set the derivative to be zero, we have 

\[
B m-B n+2 C m n-2 A=0
\]
and 
\[
\begin{aligned}
n^{\star}&=\frac{B m-2 A}{B-2 C m} \\
&= \frac{m(m^2\left(\omega_2 \sigma_{\mathrm{rag}}^2+\|\betatt\|^2 \tau_{21}\right))-2(m^3 \omega_1 \sigma^2+\|\betatt\|^2 \tau_{30} m^3+\|\betatt\|^2 \tau_2 m^2)}{(m^2\left(\omega_2 \sigma_{\mathrm{rag}}^2+\|\betatt\|^2 \tau_{21}\right)) - 2m(\left\|\beta_{\mathrm{tt}}\right\|^2\left(\tau_{22} m^2+\tau_{12} m+\tau_2\right))}\\ 
&= \frac{m\left(2 \|\betatt\|_2^2 d m+2 \|\betatt\|_2^2 d+2 \|\betatt\|_2^2 m-d m \sigma_{\text {rag}}^2+2 m \sigma^2\right)}{d\left(2 \|\betatt\|_2^2 m^2-2 \|\betatt\|_2^2 m+2 \|\betatt\|_2^2-m \sigma_{\text{rag}}^2\right)} \\ 
&\leq \frac{md\left(2 \|\betatt\|_2^2 d m-d m \sigma_{\text {rag}}^2+2 m \sigma^2\right)}{d^2\left(2 \|\betatt\|_2^2 m^2-2 \|\betatt\|_2^2 m+2 \|\betatt\|_2^2-m \sigma_{\text{rag}}^2\right)} \\ 
&= \calO \left( \frac{md\left(2 \|\betatt\|_2^2 d m-d m \sigma_{\text {rag}}^2+2 m \sigma^2\right)}{d^2\left(2 \|\betatt\|_2^2 m^2-m \sigma_{\text{rag}}^2\right)} \right) \\  
&= \calO\left(\frac{m \left(d^2 \|\betatt\|_2^2 + d \sigma^2- d^2 \sigma_{\text{rag}}^2\right)}{m d^2 \|\betatt\|_2^2 - d^2\sigma_{\text{rag}}^2}\right) \\
&= \calO\left(\frac{d\|\betatt\|_2^2 + \sigma^2-d\sigma_\rag^2 }{d\|\betatt\|_2^2}\right)
\end{aligned}
\]
where the third step follows from upper bounding the numerator, and the fourth step follows from lower bounding the denominator.
\paragraph{$n^*$ as Global Minimizer}
Now, we will show that the stationary point is the global minimizer. The second order derivative is give as: 
\begin{equation}
\begin{aligned}
\frac{\partial\left(\calL_{\text{tt+rag}}(\bW^*)\right)}{\partial n} 
& =\frac{2\left(C m^2-2 C m n-2 B m+B n+3 A\right)}{(m+n)^4}
\end{aligned}
\end{equation}
Plug in $B m-B n^*+2 C m n^*-2 A=0$, we have 
\begin{equation}
\begin{aligned}
\frac{\partial\left(\calL_{\text{tt+rag}}(\bW^*)\right)}{\partial n}  |_{n = n^*}
& = \frac{2\left(C m^2-A\right)}{(m-n)(m+n)^3}  \geq 0
\end{aligned}
\end{equation}
Since $n^* = \calO(1)$, we have $m > n^*$ for large $m$. Also, we have $Cm^2 > A$ for large $m$, thus we have $\frac{\partial\left(\calL_{\text{tt+rag}}(\bW^*)\right)}{\partial n}  |_{n = n^*} \geq 0$, and $n^*$ is the local minimum.
Now, we check the first order derivative of $n \geq n^*$, 
\[
\begin{aligned}
B m-B n+2 C m n-2 A  &=  B m-B n+2 C m n-2 A - (B m-B n^*+2 C m n^*-2 A) \\ 
&= -B (n-n^*) + 2Cm(n-n^*)  \geq 0 \\ 
\end{aligned}
\]
where it follows from $B \leq  0, C \geq 0$. Similarly, we can show that $B m-B n+2 C m n-2 A \leq 0, \quad \forall n \leq n^*$. Thus, we have $n^*$ to be the global minimum of the loss.
\paragraph{Improvement from $n^*$}
Here, we plug in $n = n^*$ and $n=0$ into~\cref{eq:loss_approx}. We have 
\begin{equation}
\begin{aligned}
\calL_{\text{tt+rag}} |_{n =n^*}(\bW^*) &= \frac{A+Bn^*+C(n^*)^2}{(m+n^*)^2(m+d+1)^2}\\
\calL_{\text{tt+rag}} |_{n=0}(\bW^*) &= \frac{A}{m^2(m+d+1)^2}\\
\end{aligned}
\end{equation}
Then, the improvement is give as 
\begin{equation}
\begin{aligned}
\calL_{\text{tt+rag}} |_{n=0}(\bW^*) - \calL_{\text{tt+rag}} |_{n =n^*}(\bW^*) &= \frac{A(m+n^*)^2 - m^2(A+Bn^*+C(n^*)^2)}{m^2(m+n^*)^2(m+d+1)^2}  \\ 
&= \frac{(n^*)^2(2 C m-B)}{2 m^2\left(m+n^*\right)(m+d+1)^2} \\ 
&= \calO\left(\frac{Cm}{m^5}  \right)\\
&= \calO\left(\frac{m^2d^2 \|\betatt\|_2^2m}{m^5} \right)\\
&= \calO\left(\frac{1}{m^2}\right)
\end{aligned}
\end{equation} 
where the second step step follows from $B m-B n^*+2 C m n^*-2 A=0$ and the third step follows from $n^* = \calO(1)$, and the four step follows from $B \leq 0$ and $|B| = \calO(C)$. It finishes the proof.
\end{proof}

\subsection{Non-Uniform Retrieval Noise}


Now, we proceed to the proof for non-uniform retrieval noise.

\subsubsection{Distance-Proportional Noise}
\begin{theorem*}[Restatement of~\cref{thm: bound_proportion_noise}]
Under Assumption~\ref{ass:gaussian_offset}, \ref{ass:data}, \ref{ass:proportional}, the population loss is given as:
\[
\hat{\err}_{\text{variance}}(\bW) = m\sigma^2 \tr(\bW^\T \bW) + \sum_{i=1}^n \gamma_1 \delta_i^2 [(1+\delta_i^2) \tr(
\bW^\T \bW
)+ \tr(\bW^2) + \tr(\bW)^2] 
\]

If the variance of the retrieval distance follows power law, i.e. $\exists \gamma_2 > 0, q \geq 0$ s.t. $ \delta_i^2 = \gamma_2 i^q $, then
\begin{equation}
\begin{aligned}
\hat{\err}_{\text{bias}}(\bW^*)&=
\calO \left(\err_\bias(\bW^*) + \|\betatt\|_2^2 \left[\frac{ dn^{2q+1} + n^{2q+2}}{(m+n)^2}\right]\right)
\end{aligned}
\end{equation}

and 
\begin{equation}
\begin{aligned}
\hat{\err}_{\text{variance}}(\bW^*)&= \calO\left(
\frac{dm\sigma^2 + d(n^{2q+1})\sigma^2}{(m+n)^2}
\right) = \begin{cases}
\calO\left(dn^{2q-1}\sigma^2\right)  & \text{ if $n\rightarrow \infty$, $q \leq 1/2$ } \\ 
\text{diverges} & \text{ if $n\rightarrow \infty$, $q>1/2$} \\ 
\end{cases} 
\end{aligned}
\end{equation}
\end{theorem*}

\begin{proof}
We first write down the error explicitly similar to \cref{eq:pop_loss_breakdown}.
\begin{equation*}
\begin{aligned}
y_q - \bx_q^\T\bW \bX^\T\by =\bx_q^\T\left(I-\bW \bG \right) \betatt-\bx_q^\T \bW \bX^\T \beps + \epsilon_q
\end{aligned}
\end{equation*}
And we can break down the population loss as 

\begin{equation}
\hat{\calL}_{\text{tt+rag}}(\bW) =\Exp_{(\bx_q, y_q),(\bX, \by),\beps, \br} \left(\bx_q^\T\left(I-\bW \bG \right) \betatt\right)^2 +  \left(\bx_q^\T \bW \bX^\T \beps\right)^2+\sigma^2 
\end{equation}
\paragraph{Variance-Induced Error}
\begin{equation}
\begin{aligned}
\hat{\err}_{\text{variance}}(\bW) &= \Exp (\bx_q^\T \bW \bX^\T \beps)^2 \\ 
&= \sum_{i,j=1}^{m+n}(\bx_i^\T \bW^\T \bx_q)(\bx_q^\T \bW \bx_j) \Exp(\epsilon_i \epsilon_j)
\end{aligned}
\end{equation}
Because the noise are independent and zero-mean, we have 
\[
\Exp[\epsilon_j \epsilon_j] = \begin{cases}
\sigma^2, & i = j \le m \\
\sigma^2_{\text{rag}, i}, & i = j > m \\
0, & i \ne j
\end{cases}
\]

Then, 

\[
\text{LHS} = \sum_{i=1}^m \sigma^2 \Exp[(\bx_q^\T \bW \bx_i)^2] + \sum_{i=m+1}^{m+n}\sigma^2_{\text{rag}, i-m} \cdot \Exp[\bx_q^\T \bW (\bx_q+\br_{i-m})^2]
\]
Thus, the ICL contribution remains the same as \cref{thm:pop-loss-rag}, i.e. 
\[ \sum_{i=1}^m \sigma^2 \Exp[(\bx_q^\T \bW \bx_i)^2] = m\sigma^2 \tr(\bW^\T \bW)\]
To compute the RAG contribution, we evaluate the formula similar to  \cref{eq:pop_loss_variance_rag}.
\begin{equation}
\begin{aligned}
\Exp\left[\left(\bx_q^\top \bW (\bx_q + \br_i)\right)^2\right]
&= \Exp[(\bx_q^\top \bW \bx_q)^2] + \Exp[(\bx_q^\top \bW \br_i)^2] + 2 \Exp[\bx_q^\top \bW \bx_q \cdot \bx_q^\top \bW \br_i] \\
&= \tr(\bW^\T \bW) + \tr(\bW^2) + \delta^2_i \tr(\bW^\T \bW) + \tr(\bW)^2
\end{aligned}
\end{equation}
And thus, the RAG error contribution is
\[
\sum_{i=m+1}^{m+n}\sigma^2_{\text{rag}, i-m} \cdot \Exp[\bx_q^\T \bW (\bx_q+\br_{i-m})^2] = \sum_{i=1}^n \sigma_{\text{rag}, i}^2 [(1+\delta_i^2) \tr(
\bW^\T \bW
)+ \tr(\bW^2) + \tr(\bW)^2] 
\]

Plug in $\sigma_{\text{rag}, i}^2 = \gamma_1\delta_i^2$, and combining all terms together, we have 
\[
\hat{\err}_{\text{variance}}(\bW) = m\sigma^2 \tr(\bW^\T \bW) + \sum_{i=1}^n \gamma_1 \delta_i^2 [(1+\delta_i^2) \tr(
\bW^\T \bW
)+ \tr(\bW^2) + \tr(\bW)^2] 
\]

Now, if we further assume $\delta_i^2 = \gamma_2 i^q$, and plug in the value of  
\begin{equation*} 
\begin{aligned}
\hat{\err}_{\text{variance}}(\bW^*) &= m\sigma^2 \tr\left((\bW^*)^\T \bW^*\right) + \sum_{i=1}^n \gamma_1 \gamma_2 i^q[(1+\gamma_2 i^q)\tr\left((\bW^*)^\T \bW^*\right)+ \tr\left((\bW^*)^2\right) + \tr(\bW^*)^2 ] \\ 
&= \frac{m^2}{(m+d+1)^2(m+n)^2} \left[  d m \sigma^2+\gamma_1 \gamma_2 \left[\left(2 d+d^2\right) \sum_{i=1}^n i^q\sigma^2+d \gamma_2 \sum_{i=1}^n i^{2 q}\sigma^2\right] \right] \\
&= \frac{m^2}{(m+d+1)^2(m+n)^2} \left[  d m \sigma^2+\gamma_1 \gamma_2\sigma^2 \left[\left(2 d+d^2\right) \calO\left(\frac{n^{q+1}}{q+1}+\frac{n^q}{2}\right)+d \gamma_2 \calO\left(\frac{n^{2q+1}}{2q+1} + \frac{n^{2q}}{2} \right)\right] \right] \\ 
&= \calO\left(
\frac{dm\sigma^2 + d(n^{2q+1})}{(m+n)^2}
\right) \\ 
&= \begin{cases}
\calO\left(dn^{2q-1}\sigma^2\right)  & \text{ if $n\rightarrow \infty$, $q < 1/2$ } \\ 
\calO\left(d\sigma^2\right) & \text{ if $n\rightarrow \infty$, $q = 1/2$} \\ 
\text{diverges} & \text{ if $n\rightarrow \infty$, $q>1/2$} \\ 
\end{cases}
\end{aligned}
\end{equation*}
where the second step follows from the Euler–Maclaurin expansion of the power sum. 

\paragraph{Bias-Induced Error}
From~\cref{eq:M4}, we note that 
\[
\err_{\bias}(\bW) = \betatt^\T \left[M_1 -M_2-M_3 + M_{41} + \sum_{i=1}^n (M_{42}+M_{42}^\T) + \sum_{i=1}^n M_{43} + \sum_{i\neq j, i,j \in [n]}M_{44} \right] \betatt
\]
Specifically, 
\begin{equation}
\begin{aligned}
\Exp_{\bx_q, \bX}\left[\left(I - \bW \bG \right)^\T \bx_q \bx_q^\T \left(I - \bW\bG\right)\right] &= \Exp_{\bx_q, \bX} \left(I-\bG \bW^\T\right) \bx_q \bx_q^\T(I-\bW \bG) \\ 
&=\underbrace{\Exp\bx_q \bx_q^\T}_{:= M_1}-\underbrace{\Exp\bx_q \bx_q^\T \bW \bG}_{:= M_2}-\underbrace{\Exp\bG \bW^\T \bx_q \bx_q^\T}_{:= M_3}+\underbrace{\Exp\bG \bW^\T \bx_q \bx_q^\T \bW \bG}_{:= M_4}
\end{aligned}
\end{equation}
To avoid the repeated computation, we will highlight the calculation that involves $\delta_i$, omit some calculation steps given in the standard case and discuss its bound after allowing for non-uniform offset. We will only compute $\delta^2_i$-involving term and use $\dots$ to denote the rest terms, since we assume $\delta^2 \ll 1$ in proving~\cref{thm:pop-loss-rag}. The final bound will be given as 
\[
\hat{\err}_{\bias}(\bW^*) = \err_{\bias}(\bW^*)  + \delta^2\text{-involved terms}
\]

$
M_1=\Exp\left[\bx_q \bx_q^\T\right]=I
$ and remains the same. Let $s_\delta := \sum_i \delta_i^2, S_\delta := \sum_i (\delta_i^2)^2$. 

Then, we expand out the terms in $M_2$: 

\begin{equation}
\begin{aligned}
M_2 = \Exp_{\bx_q, \br} \bx_q \bx_q^\T \bW \bG &= \left(\Exp_{\bx_q, \br} \bx_q \bx_q^\T\right) \bW \bG_0 + \Exp_{\bx_q, \br} \bx_q \bx_q^\T \bW \sum_{i=1}^n (\bx_q + \br_i)(\bx_q + \br_i)^\T  \\ 
&= \bW \bG_0 + \Exp_{\bx_q, \br} \bx_q \bx_q^\T \bW \sum_{i=1}^n (\bx_q \bx_q^\T + \br_i\br_i^\T)  \\ 
&= \bW \bG_0 + \Exp_{\bx_q, \br} \bx_q \bx_q^\T \bW \sum_{i=1}^n (\bx_q \bx_q^\T + \delta^2_i I)  \\ 
&= \dots  + s_\delta \bW
\end{aligned}
\end{equation}
Similarly, $M_3 = M_2^\T =\dots + s_\delta \bW^\T$. Now, we perform similar expansion for $M_4$.

First, we note that $M_{41}= \Exp_{\bx_q, \bX} [\bG_0\bW^\T \bx_q\bx_q^\T \bW\bG_0]$ is independent of $\delta_i^2$.

\begin{equation}
\begin{aligned}
\sum_{i\in[n]} M_{42} &:= \sum_{i\in[n]} \Exp_{\bx_q, \bX}\bG_0 \bW^\T \bx_q\bx_q^\T \bW \bG_i \\
&= \sum_{i\in[n]} \Exp_{\bx_q, \bX} \bG_0 \bW^\T \bx_q\bx_q^\T \bW \left( \bx_q + \br_i\right)\left(\bx_q + \br_i \right)^\T \\
&= \sum_{i\in[n]} \Exp_{\bx_q, \bX} \bG_0 \bW^\T \bx_q\bx_q^\T \bW \left( \bx_q\bx_q^\T + \br_i\br_i^\T\right) \\
&= \sum_{i\in[n]} \Exp_{\bx_q, \bX} \bG_0 \bW^\T \left(\bW+\bW^\T  + \tr(\bW) +  \bW\delta^2 \right)\\
&= \sum_{i\in[n]}  m \left(\bW^\T  \bW+\bW^\T \bW^\T  + \tr(\bW ) \bW^\T  + \delta^2  \bW^\T \bW \right) \\
&= \dots + m s_{\delta} \bW^\T \bW \\
\end{aligned}
\end{equation}
Following the derivation of the 6th-order and 4th-order moments as in~\cref{lem:6order_gaussian} and~\cref{lem:mixed-4th-order}, we have
\begin{equation}
\begin{aligned}
\sum_{i\in[n]}  M_{43} &:= \sum_{i\in[n]} \Exp_{\bx_q, \bX, \br_i}\bG_i \bW^\T \bx_q \bx_q^\T\bW \bG_i \\ 
&= \left.\begin{aligned}
    & 2\left(\bW^2 + (\bW^2)^\T  + \bW^\T \bW+ \bW\bW^\T + \tr\left(\bW\right)(\bW + \bW^\T)\right)  \\ 
    & + \tr\left(\bW\right)^2 I + \tr\left(\bW^2\right)I + \tr \left(\bW^\T \bW \right)I \\ 
\end{aligned} \right\} \text{0th-order in $\br_i$, \cref{lem:6order_gaussian}}\\ 
&\quad+ \underbrace{2\delta_i^4 \bW^\T\bW + \delta_i^4 \tr(\bW^\T \bW)I}_{\text{4th-order in $\br_i$, \cref{eq:M4_4order_term}}} \\ 
&\quad+ \underbrace{\delta_i^2 \left[\tr\left(\bW \right)\left(\bW^\T + \bW \right) +\bW^2+(\bW^2)^\T+ 2\bW^\T \bW  \right]}_{\text{\cref{eq:M4_2order_term4} and its transpose}} \\ 
&\quad+ \underbrace{\left(\tr\left(\bW^2 \right) + \tr\left( \bW^\T \bW\right) + \tr\left(\bW \right)^2\right)\delta_i^2 I}_{\text{\cref{eq:M4_2order_term1}}} \\ 
&\quad+ \underbrace{ 2 \delta_i^2  \bW\bW^\T+\delta_i^2  \tr(\bW^\T\bW )I}_{\text{\cref{eq:M4_2order_term2}}} \\ 
&\quad+ \underbrace{\delta_i^2 \left[\tr\left(\bW \right)\left(\bW^\T + \bW \right) +\bW^2+(\bW^2)^\T+ 2\bW^\T \bW\right]}_{\text{\cref{eq:M4_2order_term3} and its transpose}} \\
&= \sum_{i \in [n]}(2+2\delta_i^2) \left[\tr\left(\bW \right)\left(\bW^\T + \bW \right) +\bW^2+(\bW^2)^\T\right]  \\ 
&\quad + \sum_{i \in [n]}(2+4\delta_i^2)\bW^\T \bW + \sum_{i \in [n]}2\bW\bW^\T \\ 
&\quad + \sum_{i \in [n]}(1+\delta_i^2)\left[\tr\left(\bW\right)^2 I + \tr\left(\bW^2\right)I + \tr \left(\bW^\T \bW \right)I \right]\\ 
&\quad+ \sum_{i \in [n]}\left( 2\delta_i^4 \bW^\T\bW + \delta_i^4 \tr(\bW^\T \bW)I +  2 \delta_i^2  \bW\bW^\T+\delta_i^2  \tr(\bW^\T\bW )I\right) \\
&= \dots +  2s_\delta \left[\tr\left(\bW \right)\left(\bW^\T + \bW \right) +\bW^2+(\bW^2)^\T \right]  \\
&\quad +(4s_\delta+2S_\delta)\bW^\T\bW + 2s_\delta\bW\bW^\T \\ 
&\quad+s_\delta\left(\tr\left(\bW\right)^2 + \tr\left(\bW^2\right)\right)I + (2s_\delta+S_\delta) \tr(\bW^\T\bW)I \\ 
\end{aligned}
\end{equation}

Also, we expand the cross-term out for $\forall i,j \in [n], i\neq j$: 
\begin{equation}
\begin{aligned}
\sum_{i\neq j} M_{44}&:=\sum_{i\neq j} \Exp\bG_i \bW^\T \bx_q \bx_q^\T\bW \bG_j \\ 
&= \sum_{i\neq j} \left(\bx_q \bx_q^\T \bW^\T \bx_q \bx_q^\T \bW \bx_q \bx_q^\T + \br_i\br_i^\T \bW^\T \bx_q \bx_q^\T \bW \bx_q\bx_q^\T\right) \\ 
&\quad + \sum_{i\neq j}\left( \bx_q\bx_q^\T \bW^\T \bx_q \bx_q^\T \bW \br_j \br_j^\T + \br_i \br_i^\T \bW^\T \bx_q \bx_q^\T \bW \br_j \br_j^\T \right) \\ 
&= \dots + \sum_{i\neq j} \delta^2_i\left(\bW^2  + \bW^\T \bW+ \tr\left(\bW\right)\bW  \right)  \\ 
&\quad + \sum_{i\neq j}\delta^2_i\left(\left(\bW^2\right)^\T + \bW^\T \bW+ \tr\left(\bW\right) \bW^\T \right)  \\ 
&\quad + \sum_{i\neq j} \delta^2_i \delta_j^2 \bW^\T \bW\\ 
\end{aligned}
\end{equation}
In the non-uniform noise scenario, 4th-order term in $\delta_i$ will dominate the 2nd-order term in $\delta_i$. Thus, we will plug $\delta_i^2 =\gamma_2 i^q, \bW^* = \frac{m}{(m+d+1)(m+n)}$ into $\err_\bias$: 
\begin{align*}
\err_{\text{bias}}(\bW^*)
&= \err_\bias(\bW^*) + \calO\left(
\betatt^\T \left[ 2 \sum_i^n (\delta_i^2)^2 (\bW^*)^\T \bW^*  +\sum_{i\neq j, i\in [n], j\in[n]} \delta_i^2\delta_j^2 (\bW^*)^\T \bW^* +  \sum_i^n (\delta_i^2)^2 \tr((\bW^*)^\T \bW) I  \right] \betatt
\right) \\ 
&= \err_\bias(\bW^*) + \calO\left(
\betatt^\T \left[ dn^{2q+1} (\bW^*)^\T \bW^*+ n^{2q+2} (\bW^*)^\T \bW^*\right] \betatt
\right)\\
&=\err_\bias(\bW^*) +  \calO\left(\betatt^\T \left[\frac{ dn^{2q+1} + n^{2q+2}}{(m+n)^2} \right]\betatt\right)
\end{align*}
It finishes the proof.
\end{proof}

\subsubsection{Distance-Weighted Probabilistic Noise}

\begin{theorem*}[Restatement of~\cref{thm:bound_probablistic_noise}]
Under Assumption~\ref{ass:gaussian_offset}, \ref{ass:data}, \ref{ass:probablistic}, then $\tilde{\err}_\bias(\bW) = \hat{\err}_\bias(\bW)$, and
\[
\hat{\err}_{\text{variance}}(\bW) = m\sigma^2 \tr(\bW^\T \bW) + \sum_{i=1}^n \left(p_i\sigma_s^2  + (1-p_i)\sigma_l^2 \right) [(1+\delta_i^2) \tr(
\bW^\T \bW
)+ \tr(\bW^2) + \tr(\bW)^2] 
\]
If the variance of the retrieval distance follows power law, i.e. $\exists \gamma_2 > 0, q \geq 0$ s.t. $ \delta_i^2 = \gamma_2 i^q $, then:

\begin{equation}
\begin{aligned}
\tilde{\err}_{\text{variance}}(\bW^*)  
& = \begin{cases}
\calO\left( c_l dn^{q-1}\sigma^2 - (c_l-c_s) \sigma^2 d n^{q-1-q \tilde{q}} \right) & \text{ if $n\rightarrow \infty$, $q \leq 1$} \\ 
\text{diverges} & \text{ if $n\rightarrow \infty$, $q>1$} \\  
\end{cases}
\end{aligned}
\end{equation}

\end{theorem*}

\begin{proof}
First, we note that $\tilde{\err}_\bias(\bW) = \hat{\err}_\bias(\bW)$, since both are independent of $\sigma_\text{rag}^2$ and depend on the same set of $\forall i, \; \delta_i^2$.

We write down error explicitly similar to \cref{eq:pop_loss_breakdown} and break down the population loss as:
\begin{equation}
\tilde{\calL}_{\text{tt+rag}}(\bW) =\Exp_{(\bx_q, y_q),(\bX, \by),\beps, \br} \left(\bx_q^\T\left(I-\bW \bG \right) \betatt\right)^2 +  \left(\bx_q^\T \bW \bX^\T \beps\right)^2+\sigma^2 
\end{equation}
We note that $\tilde{\err}_{\text{bias}}(\bW) = \err_{\text{bias}}(\bW)$, since the error from bias does not depend on the sample complexity. 
\begin{equation}
\begin{aligned}
\tilde{\err}_{\text{variance}}(\bW) &= \Exp (\bx_q^\T \bW \bX^\T \beps)^2 \\ 
&= \sum_{i,j=1}^{m+n}(\bx_i^\T \bW^\T \bx_q)(\bx_q^\T \bW \bx_j) \Exp(\epsilon_i \epsilon_j)
\end{aligned}
\end{equation}
Because the noise are independent and zero-mean, we have 
\[
\Exp[\epsilon_j \epsilon_j] = \begin{cases}
\sigma^2, & i = j \leq m\\
\sigma^2_s, & i = j > m, \text{ w.p. } p\\
\sigma^2_l, & i = j > m, \text{ w.p. } 1-p\\
0, & i \ne j
\end{cases}
\]
Thus, the ICL contribution remains the same as \cref{thm:pop-loss-rag}, i.e. 
\[ \sum_{i=1}^m \sigma^2 \Exp[(\bx_q^\T \bW \bx_i)^2] = m\sigma^2 \tr(\bW^\T \bW)\]
To compute the RAG contribution, we evaluate the formula similar to  \cref{eq:pop_loss_variance_rag}.
\begin{equation}
\begin{aligned}
\Exp\left[\left(\bx_q^\top \bW (\bx_q + \br_i)\right)^2\right]
&= \Exp[(\bx_q^\top \bW \bx_q)^2] + \Exp[(\bx_q^\top \bW \br_i)^2] + 2 \Exp[\bx_q^\top \bW \bx_q \cdot \bx_q^\top \bW \br_i] \\
&= \tr(\bW^\T \bW) + \tr(\bW^2) + \delta^2_i \tr(\bW^\T \bW) + \tr(\bW)^2
\end{aligned}
\end{equation}
And thus, the RAG error contribution is
\[
\sum_{i=1}^n \left(p_i\sigma_s^2  + (1-p_i)\sigma_l^2 \right) [(1+\delta_i^2) \tr(
\bW^\T \bW
)+ \tr(\bW^2) + \tr(\bW)^2] 
\]

Plug in $\sigma_{\text{rag}, i}^2 = \gamma_1\delta_i^2$, and combining all terms together, we have 
\[
\hat{\err}_{\text{variance}}(\bW) = m\sigma^2 \tr(\bW^\T \bW) + \sum_{i=1}^n \left(p_i\sigma_s^2  + (1-p_i)\sigma_l^2 \right) [(1+\delta_i^2) \tr(
\bW^\T \bW
)+ \tr(\bW^2) + \tr(\bW)^2] 
\]

Now we further assume $p_i = (1+\delta_i^2)^{-\tilde{q}}, \; \tilde{q} \geq 0$, and plug in the value of $\bW^*$. Let $B := \frac{m^2}{(m+d+1)^2(m+n)^2}$, 
\begin{equation*} 
\begin{aligned}
\tilde{\err}_{\text{variance}}(\bW^*) &= m\sigma^2 \tr(\bW^\T \bW) + \sum_{i=1}^n \left(p_i\sigma_s^2  + (1-p_i)\sigma_l^2 \right) [(1+\delta_i^2) \tr(
\bW^\T \bW
)+ \tr(\bW^2) + \tr(\bW)^2] \\ 
&= B \left[  d m \sigma^2+ \sum_{i=1}^{n}\left( c_l\sigma^2 - (1+\delta_i^2)^{-\tilde{q}}(c_l -c_s)\sigma^2 \right)\left[ (1+ \delta_i^2) \cdot d+  d + d^2\right] \right] \\
&\approx  B\left[  d m \sigma^2+ c_l \sigma^2\sum_{i=1}^n \left(d\delta_i^2 + d^2\right) - (c_l-c_s) \sigma^2\sum_{i=1}^n \left( d(1+\delta_i^2)^{1-\tilde{q}} + d^2 (1+\delta_i^2)^{-\tilde{q}}\right)\right]  \\
&\approx  B \left[  d m \sigma^2+ c_l \sigma^2\sum_{i=1}^n d\delta_i^2 - (c_l-c_s) \sigma^2\sum_{i=1}^n d(1+\delta_i^2)^{1-\tilde{q}} \right]  \\
&\approx 
\begin{cases}
B \left[  d m \sigma^2+ c_l \sigma^2d n^{q+1} - (c_l-c_s) \sigma^2 d\log(n)\right]  & \text{ if $\tilde{q} = 1+1/q$}\\
B \left[  d m \sigma^2+ c_l \sigma^2d n^{q+1} - (c_l-c_s) \sigma^2 d n^{1+q -q \tilde{q}} \right]  & \text{ else}\\ 
\end{cases} \\ 
\end{aligned}
\end{equation*}
where the second line follows from omitting the lower order term.

If $\tilde{q} = 1 + 1/q$, we note that the middle term will dominate the error. And combining all cases, we could obtain

\[
\begin{aligned}
\tilde{\err}_{\text{variance}}(\bW^*)  
& = \begin{cases}
\calO\left( c_l dn^{q-1}\sigma^2 - (c_l-c_s) \sigma^2 d n^{q-1-q \tilde{q}} \right) & \text{ if $n\rightarrow \infty$, $q \leq 1$} \\ 
\text{diverges} & \text{ if $n\rightarrow \infty$, $q>1$} \\  
\calO\left( c_l dn^{q-1}\sigma^2 + (c_l - c_s)d^2 \frac{\log n}{n^2} \sigma^2\right) & \text{ if $n\rightarrow \infty$, $\tilde{q} = 1 + 1/q$} \\ 
\end{cases}
\end{aligned}
\]

\end{proof}

\end{document}